\pdfoutput=1
\documentclass[11pt]{article}
\usepackage[final]{acl}
\usepackage{times}

\usepackage[table]{xcolor}
\usepackage{colortbl}
\usepackage{booktabs,multirow}
\usepackage{siunitx}
\usepackage{adjustbox}
\usepackage{algorithmic}
\usepackage{latexsym}
\usepackage[T1]{fontenc}
\usepackage[utf8]{inputenc}
\usepackage{microtype}
\usepackage{inconsolata}
\usepackage{graphicx}
\usepackage{booktabs}
\usepackage{amsfonts}
\usepackage{nicefrac}
\usepackage{microtype}
\usepackage{comment}
\usepackage{subcaption}
\usepackage{caption}
\usepackage{float}
\usepackage{amsthm, amsmath, amsfonts, amssymb}
\usepackage{mathtools}
\usepackage{extarrows}
\usepackage{multirow}
\usepackage{enumitem}
\usepackage{amsmath,amsfonts,bm}

\def\Figref#1{Figure~\ref{#1}}
\def\Tabref#1{Table~\ref{#1}}

\def\Secref#1{Section~\ref{#1}}
\def\eqref#1{equation~\ref{#1}}
\def\Eqref#1{Equation~\ref{#1}}
\def\1{\bm{1}}

\DeclareMathAlphabet{\mathsfit}{\encodingdefault}{\sfdefault}{m}{sl}
\SetMathAlphabet{\mathsfit}{bold}{\encodingdefault}{\sfdefault}{bx}{n}

\usepackage{wrapfig}
\usepackage{lipsum}
\theoremstyle{plain}

\theoremstyle{definition}

\theoremstyle{remark}

\usepackage[normalem]{ulem}
\usepackage{longtable}
\usepackage{makecell}
\usepackage{arydshln}
\usepackage{adjustbox}
\usepackage{threeparttable}
\usepackage{xurl}

\usepackage{tcolorbox}

\definecolor{forward}{RGB}{84, 130, 53}
\definecolor{inverse}{RGB}{47, 85, 151}
\definecolor{resist}{RGB}{128, 0, 128}
\definecolor{rebound}{RGB}{133, 19, 33}

\definecolor{def}{RGB}{119, 228, 200}
\definecolor{thm}{RGB}{69, 53, 193}

\newtcolorbox{thmbox}[1][]{colback=thm!5!white,colframe=thm!60!black,boxsep=-4pt,grow to left by=4pt,left=10pt,grow to right by=4pt,right=10pt,top=10pt,bottom=10pt,#1}
\newtcolorbox{defbox}[1][]{colback=def!5!white,colframe=def!60!black,boxsep=-4pt,grow to left by=4pt,left=10pt,grow to right by=4pt,right=10pt,top=10pt,bottom=10pt,#1}

\newcommand{\Hglob}{\widehat{\mathcal{H}}}

\title{Revisiting Entropy Regularization: Adaptive Coefficient \\ Unlocks Its Potential for LLM Reinforcement Learning}

\author{
\textbf{Xiaoyun Zhang}$^{1, 3, \dagger}$\quad
\textbf{Xiaojian Yuan}$^{2, \dagger}$\quad
\textbf{Di Huang}$^{1}$\quad
\textbf{Wang You}$^{4}$\\
\textbf{Chen Hu}$^{4}$\quad
\textbf{Jingqing Ruan}$^{3}$\quad
\textbf{Ai Jian}$^{4}$\quad
\textbf{Kejiang Chen}$^{2}$\quad
\textbf{Xing Hu}$^{1, \ast}$\\
$^1$State Key Lab of Processors, Institute of Computing Technology, CAS\\
$^2$University of Science and Technology of China, $^3$University of Chinese Academy of Sciences\\
$^4$StepFun Inc\\
\texttt{zhangxiaoyun24@mails.ucas.ac.cn} \quad \texttt{xjyuan@mail.ustc.edu.cn}
}

\begin{document}
\maketitle

\renewcommand{\thefootnote}{\fnsymbol{footnote}}
\footnotetext[2]{Xiaoyun Zhang and Xiaojian Yuan contributed equally to this work.}
\footnotetext[1]{Corresponding author.}
% \footnotetext[3]{Under review at ACL Rolling Review, October 2025.}

\renewcommand{\thefootnote}{\arabic{footnote}}

\begin{abstract}
Reasoning ability has become a defining capability of Large Language Models (LLMs), with Reinforcement Learning with Verifiable Rewards (RLVR) emerging as a key paradigm to enhance it. However, RLVR training often suffers from policy entropy collapse, where the policy becomes overly deterministic, hindering exploration and limiting reasoning performance. While entropy regularization is a common remedy, its effectiveness is highly sensitive to the fixed coefficient, making it unstable across tasks and models. In this work, we revisit entropy regularization in RLVR and argue that its potential has been largely underestimated. Our analysis shows that (i) tasks of varying difficulty demand distinct exploration intensities, and (ii) balanced exploration may require the policy entropy to be maintained within a moderate range below its initial level. Therefore, we propose Adaptive Entropy Regularization (AER) — a framework that dynamically balances exploration and exploitation via three components: difficulty-aware coefficient allocation, initial-anchored target entropy, and dynamic global coefficient adjustment. Experiments on multiple mathematical reasoning benchmarks show that AER consistently outperforms baselines, improving both reasoning accuracy and exploration capability.

\end{abstract}

\section{Introduction}
\label{sec:introduction}
Reasoning ability has become a crucial capability for Large Language Models (LLMs) to solve complex tasks in mathematics and coding. Reinforcement Learning with Verifiable Rewards (RLVR) has recently emerged as an effective paradigm to enhance this capability, driving advances in state-of-the-art models such as OpenAI-o1 and DeepSeek-R1~\citep{jaech2024openai, guo2025deepseek}. However, recent studies observe that \emph{policy entropy collapse} may pose a significant bottleneck in RLVR training~\citep{cui2025entropy, he2025skywork, cheng2025reasoning, dai2025cde}, closely tied to the long-standing exploration–exploitation dilemma~\citep{sutton1998reinforcement}. Specifically, the model's policy often converges prematurely to a narrow set of exploitative reasoning trajectories, thereby suppressing exploration of the broader solution space~\citep{chen2025pass}. This premature convergence typically manifests as a rapid decline in policy entropy during the early stages of training~\citep{yu2025dapo}, trapping the policy in local optima and leading to performance plateaus that constrain the model's overall reasoning potential~\citep{cui2025entropy}.

% \vspace{-3mm}
\begin{figure*}[t]
\centering
\includegraphics[width=0.95 \linewidth]{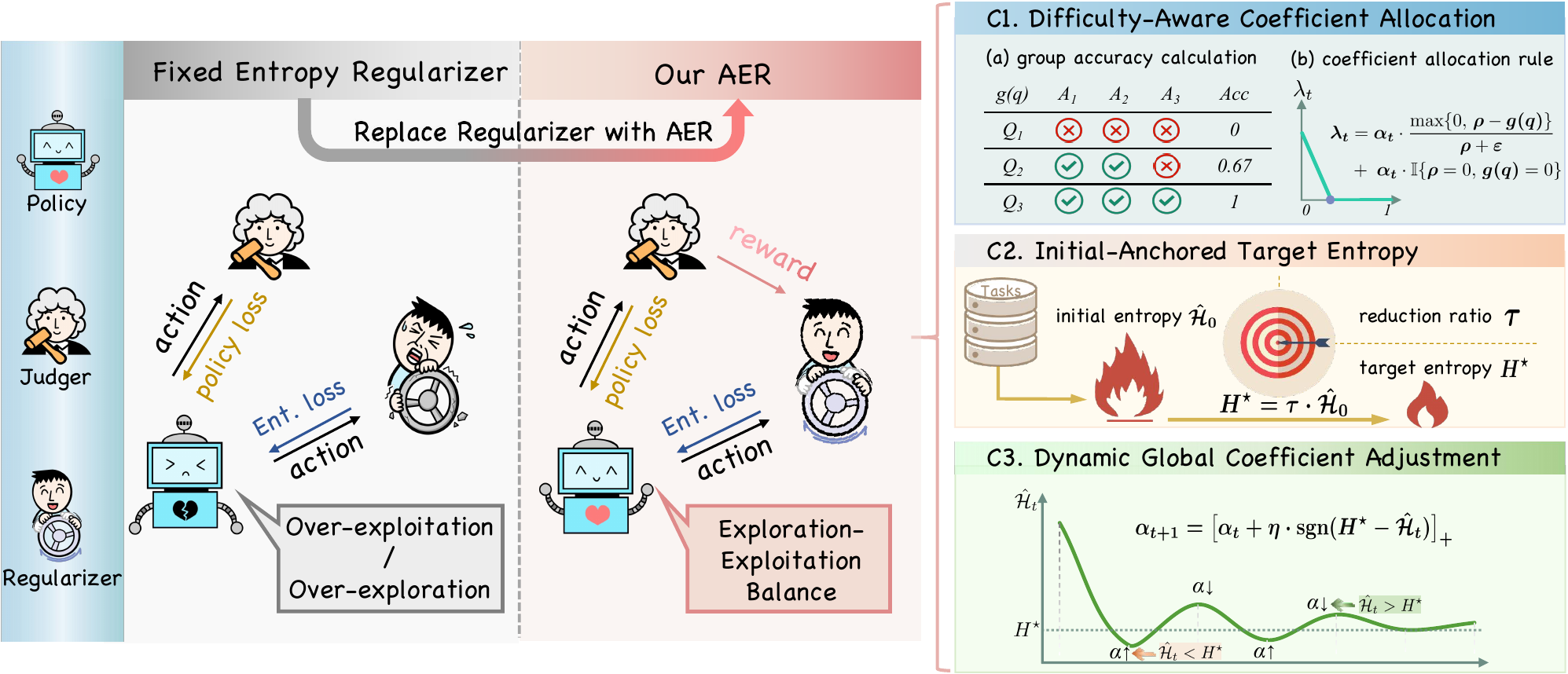}
\vspace{-2mm}
\caption{An overview of the AER framework.}
\label{fig:AER_framework}
\end{figure*}
% \vspace{-2mm}

A conventional approach in reinforcement learning to alleviate policy entropy collapse is to introduce an entropy regularization term, which explicitly penalizes overly deterministic policies and encourages exploration~\citep{schulman2017proximal}. Despite its simplicity and conceptual appeal, this technique is often omitted in recent RLVR pipelines for LLMs~\citep{yu2025dapo, hu2025open, liu2025understanding, cui2025process}, as its effectiveness is highly sensitive to the choice of the entropy coefficient. Small coefficients cannot prevent entropy collapse, whereas excessively large coefficients may induce entropy explosion~\citep{cui2025entropy, jiang2025rethinking}. Moreover, a slight change in the base model or dataset may flip the effect of a tuned coefficient from beneficial to harmful~\citep{he2025skywork}.
Intuitively, the balance between exploration (high entropy) and exploitation (low entropy) should be dynamic throughout training. Fixed coefficients struggle to deal with this evolving trade-off~\citep{he2025skywork, cui2025entropy}. This naturally raises the question:
\begin{center}
    \emph{Can we adaptively adjust the coefficient for entropy regularization during RLVR training?}
\end{center}

In this work, we revisit entropy regularization in the context of RLVR for LLMs and argue that its potential has been largely underestimated due to the limitations of fixed-coefficient designs. 
Motivated by this concern, we conduct preliminary analysis in~\Secref{sec:pre_analysis} and have two observations: (\emph{i}) tasks of different difficulty levels require distinct exploration intensities, suggesting the need for a difficulty-aware mechanism that enables sample-level control of entropy regularization; and (\emph{ii}) effective exploration during training requires maintaining the policy entropy at a specific target value below its initial entropy.

Therefore, we propose \emph{Adaptive Entropy Regularization (AER)} as shown in~\Figref{fig:AER_framework}, which dynamically balances exploration and exploitation through adaptive coefficients, including three components:
(\emph{i}) \emph{Difficulty-Aware Coefficient Allocation} estimates task difficulty relative to the current policy and assigns sample-level entropy coefficients to achieve fine-grained entropy regularization;  
(\emph{ii}) \emph{Initial-Anchored Target Entropy} adaptively determines the target entropy value based on each run’s initial entropy,  maintaining consistent relative exploration budget among different settings; and 
(\emph{iii}) \emph{Dynamic Global Coefficient Adjustment} adaptively adjusts a global scaling factor for entropy coefficients according to the current policy entropy to ensure that the policy entropy is maintained near the target entropy during training.
Together, these components form an adaptive controller that maintains policy entropy within a reasonable range, stabilizing training while retaining balanced exploration. We conducted empirical evaluations on various complex mathematical reasoning benchmarks, and AER showed consistent improvements in reasoning performance and diversity. Our contributions are summarized as follows:
\begin{itemize}[itemsep=-1pt]
    \item We conduct preliminary analysis to show that exploration should adapt to task difficulty and that balanced exploration may require maintaining the policy entropy within a moderate range below its initial level, motivating adaptive, difficulty-aware entropy regularization.
    \item We introduce the \emph{Adaptive Entropy Regularization (AER)}, a framework that can dynamically and adaptively adjust the coefficients of entropy regularization to better balance exploration and exploitation throughout training.
    \item Extensive experiments on various mathematical reasoning benchmarks demonstrate that AER consistently outperforms advanced baselines in both reasoning performance (\textit{pass@1}) and exploration capability \textit{(pass@$k$}), validating the potential of adaptive entropy regularization in RLVR training.
\end{itemize}

% AER_framework.pdf
    % \item We propose \emph{Adaptive Entropy Regularization (AER)}, a dynamic framework that adaptively balances exploration and exploitation through three components: difficulty-aware coefficient allocation, initial-anchored target entropy, and dynamic global coefficient adjustment.
\vspace{-2mm}
\begin{figure*}[t]
  \centering
  \begin{subfigure}[t]{0.31\textwidth}
    \centering
    \includegraphics[width=\linewidth]{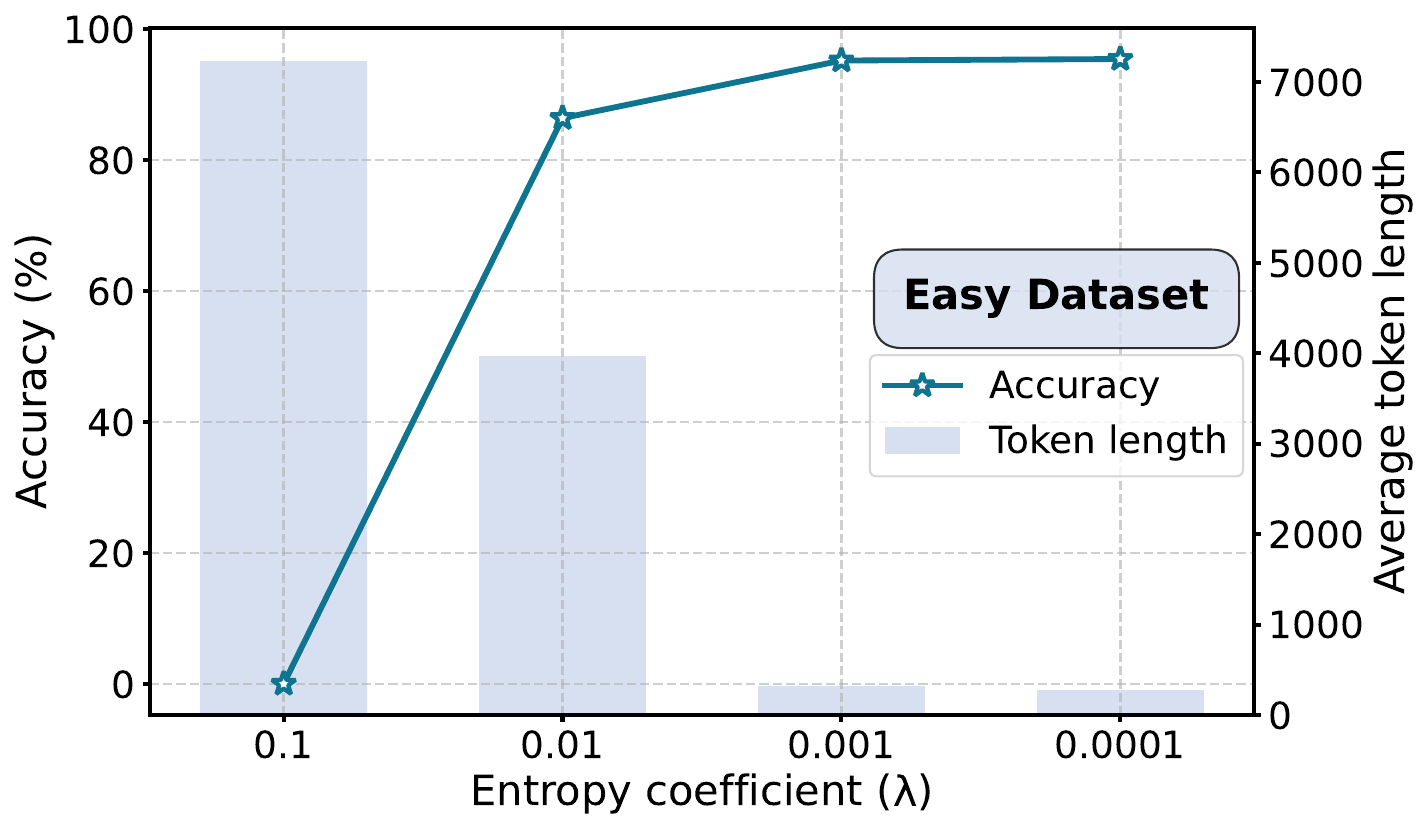}
    \caption{The impact of different coefficients on the easy dataset.}
    \label{fig:pre_exp_1a}
  \end{subfigure}\hfill
  \begin{subfigure}[t]{0.31\textwidth}
    \centering
    \includegraphics[width=\linewidth]{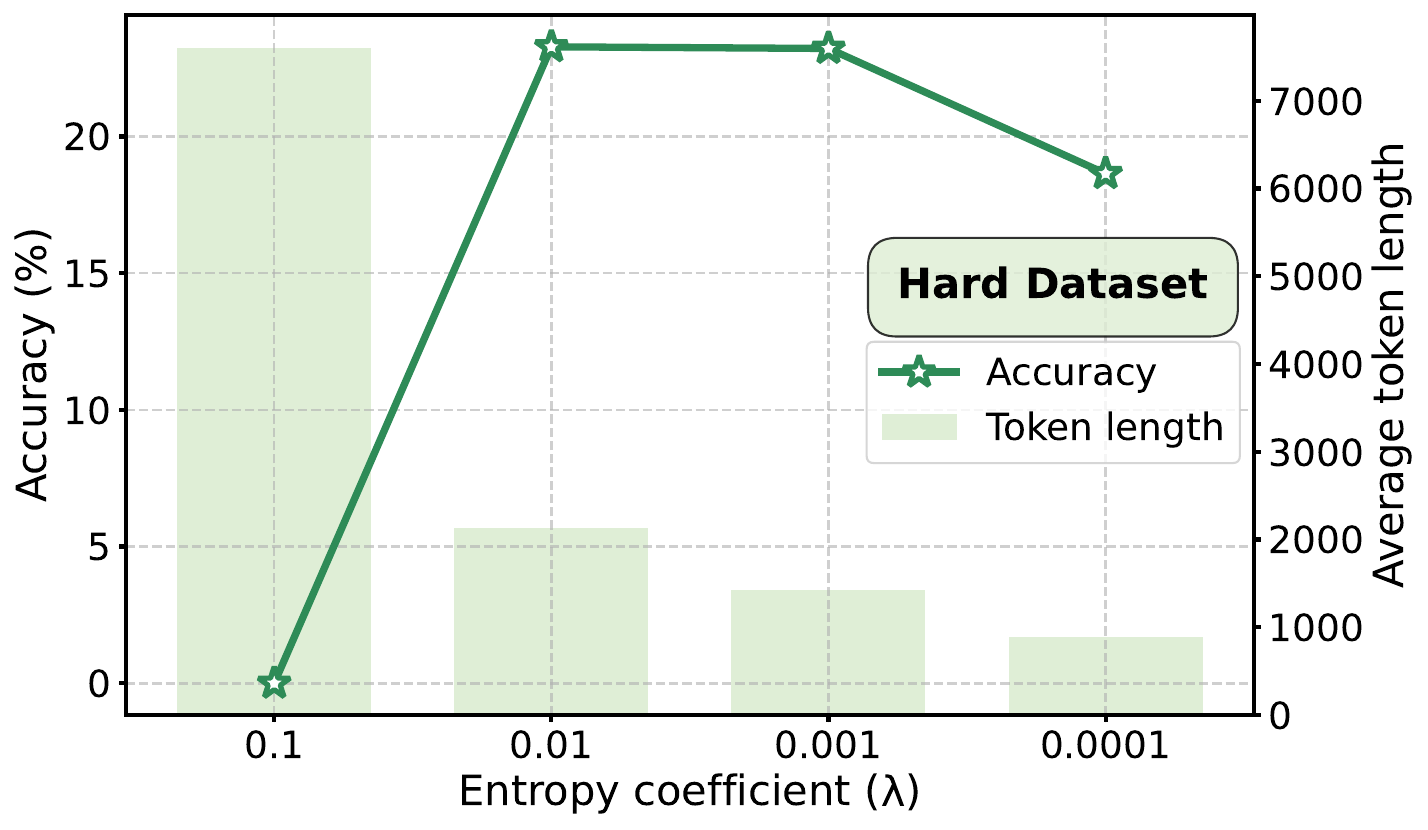}
    \caption{The impact of different coefficients on the hard dataset.}
    \label{fig:pre_exp_1b}
  \end{subfigure}\hfill
  \begin{subfigure}[t]{0.38\textwidth}
    \centering
    \includegraphics[width=\linewidth]{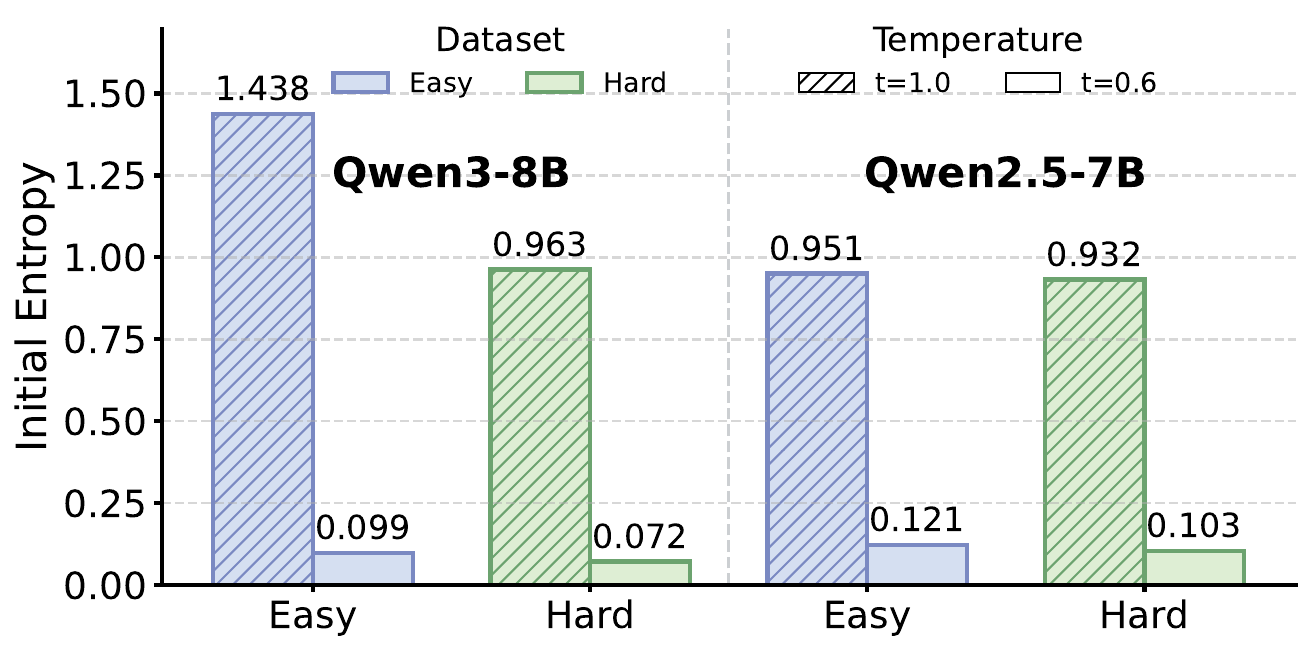}
    \caption{Initial entropy under various settings.}
    \label{fig:pre_exp_1c}
  \end{subfigure}
  \caption{\textbf{Preliminary experimental results.} (a-b) we show the effect of different entropy coefficients on test accuracy and average token length on easy and difficult datasets, respectively. (c) we demonstrate that different base models, datasets, and sampling temperatures will significantly affect the value of the initial entropy.}
  \label{fig:pre_exp}
\end{figure*}

\section{Related Work}
\label{sec:related_work}
\paragraph{Reinforcement Learning for LLMs.} Reinforcement learning is an important paradigm for training LLMs~\citep{ouyang2022training, team2023gemini, lee2024rlaif, jian2026patarmbridgingpairwisepointwise}.
Recently, reinforcement learning with verifiable rewards (RLVR) has shown remarkable success, demonstrating significant performance in complex tasks~\citep{guo2025deepseek, jaech2024openai, yang2025qwen3, he2025skywork,team2025kimi,liu2025understanding,zeng2025simplerlzoo,zhang2025reasoner, luo2026sparse}. In addition, a series of actor-only methods further reduce the resource burden and complexity of RLVR~\citep{shao2024deepseekmath, li2024remax, hu2025reinforce++, yu2025dapo, zheng2025group, li2026sesearchselfevolvingsearchagent}. However, RLVR still suffers from exploration-exploitation dilemma, which manifests as a rapid decrease in policy entropy, thus limiting the performance of LLMs~\citep{cui2025entropy, cheng2025reasoning, he2025skywork, zhang2025continue}.

\paragraph{Exploration in Reinforcement Learning.} 
Exploration is a central challenge in reinforcement learning, typically approached through theoretical analysis~\citep{cai2020provably, ishfaq2021randomized}, curiosity-driven signals~\citep{pathak2017curiosity, burdaexploration, raileanuride, henaff2022exploration}, and entropy maximization~\citep{ziebart2008maximum, toussaint2009robot}. In the context of LLMs, several studies have employed entropy as a performance indicator~\citep{cui2025entropy} or as a heuristic for advantage shaping, enhancing the rollout phase, or loss masking~\citep{wang2025beyond, cheng2025reasoning, zheng2025first, li2025staying}. Entropy regularization or KL penalty helps control policy distributions~\citep{he2025skywork, liu2025prorl}, while complementary techniques such as loss reweighting~\citep{wang2025beyond, cui2025entropy} and clip-higher~\citep{yu2025dapo} further mitigate entropy collapse. Additional strategies for promoting exploration include adjusting sampling hyperparameters~\citep{chen2025empirical}, performing self-reflection~\citep{jiang2025pag}, leveraging external verification~\citep{zha2025rl}, emphasizing high-entropy tokens via critical-token training ~\citep{wang2025beyond,li2025cure, jiang2025rethinking}, and designing custom intrinsic signals~\citep{li2025know,dai2025cde,gao2025navigate, song2025outcome}. However, the necessity of entropy regularization in RLVR remains debated, with some studies questioning its impact on exploration effectiveness~\citep{ouyang2022training,shao2024deepseekmath,hu2025open, cui2025entropy}.

\section{Preliminary Analysis}
\label{sec:pre_analysis}
Although adding explicit entropy regularization (e.g., an entropy loss term in the objective) is a straightforward, plug-and-play remedy to prevent the policy from becoming overly deterministic and thus mitigate entropy collapse, most recent works of RLVR for LLMs do not include this technique~\citep{hu2025open, liu2025understanding, cui2025process,yu2025dapo}. Yet the entropy regularization is highly sensitive to coefficients in practice, small coefficients cannot effectively prevent entropy collapse, while large coefficients will lead to entropy explosion, causing training instability or performance degradation as well~\citep{cui2025entropy}. Moreover, slight changes in the experimental setup can cause the carefully selected coefficients to have the opposite effect~\citep{he2025skywork}.

Intuitively, the degree of exploration should correlate with task difficulty: excessive exploration on easy tasks may introduce unnecessary randomness and hinder convergence, whereas difficult tasks often require stronger exploration to escape local optima and discover effective reasoning trajectories~\citep{li2025know, li2025questa, jiang2026foeforesterrorsmakes}.
To examine this intuition, we train Qwen3-4B-Base with GRPO on mathematical datasets of different difficulty levels\footnote{The easy task uses GSM8K, while the hard task consists of a mixture of AIME and AMC datasets.}, varying the strength of entropy regularization through different coefficients.
As shown in~\Figref{fig:pre_exp_1a} and~\Figref{fig:pre_exp_1b}, increasing the entropy coefficient improves test accuracy on the harder dataset, accompanied by a longer average token length—indicating that moderate promotion of exploration benefits challenging reasoning tasks. In contrast, on the easier dataset, stronger entropy regularization leads to a decline in accuracy, where excessive exploration with longer responses may prevent convergence toward concise and correct reasoning trajectories.
These results demonstrate that the optimal level of exploration varies with task difficulty, \emph{highlighting the necessity of a difficulty-aware mechanism for entropy regularization}. 
In addition, when the coefficient is set to an overly large value (i.e., $0.1$), both datasets showed a sudden drop in accuracy accompanied by a sharp increase in the average token length, indicating that \emph{excessive exploration may trigger entropy explosion, leading to instability in the training process}.

A recent study~\citep{cui2025entropy} establishes an empirical relation between policy entropy $\mathcal{H}$ and downstream performance $\mathcal{R}$, expressed as $\mathcal{R} = -a \exp{(\mathcal{H})} + b $, where $a$ and $b$ are fitting coefficients. This indicates an inherent trade-off: policy performance is ``purchased'' at the cost of entropy. 
Furthermore, policy entropy decreases monotonically without any entropy intervention~\citep{cui2025entropy}, which means that  effective exploration may require maintaining policy entropy at a "sweet spot" below its initial level to avoid entropy collapse and explosion.~\citet{he2025skywork} have similar empirical observations that they monitor policy entropy during training and preventing it from falling below a prespecified target entropy value (e.g., 0.2) from the initial entropy. However, as shown in~\Figref{fig:pre_exp_1c}, the initial entropy may vary greatly due to the differences in the base model, training data, and sampling temperature, leading to inconsistent ``exploration budget''. \emph{This inspires us to design a mechanism that adaptively determines the target entropy value based on the level of initial entropy}.

\section{Methodology}
\label{sec:method}
Building on these insights, we propose the \textbf{Adaptive Entropy Regularization (AER)} framework for RLVR.
AER estimates task difficulty with respect to the current policy and adaptively adjusts the entropy coefficient at the sample level.
It further sets the target entropy as a fraction of the initial policy entropy and dynamically adjusts a global scaling factor for coefficients to prevent the policy entropy from falling below this target, maintaining effective exploration throughout training.

\subsection{Notations and Preliminaries}
\paragraph{Formulation.}
We consider reinforcement learning with verifiable rewards (RLVR) for training LLMs. Given a question-answer pair $(q,a)$ from the dataset $\mathcal{D}$, the policy model $\pi_\theta(\cdot \mid q)$ with parameters $\theta$ generates a response $o$. A rule-based reward function $r(q,o) \in \{0,1\}$ provides binary correctness judgment based on $(q,a).$
% , where $1$ indicates correct and $0$ indicates incorrect.

\paragraph{Group Relative Policy Optimization (GRPO).}
GRPO~\citep{shao2024deepseekmath} extends proximal policy optimization~\citep{schulman2017proximal} to the group sampling setting and eliminates the need for a value network. For each question $q$, a group of $G$ candidate responses $\{o_i\}_{i=1}^G$ is sampled. 
Then, the advantage of the $i$-th response is calculated by normalizing the group-level rewards $\{ R_i \}_{i=1}^G:=\{r(q,o_1),\dots,r(q,o_G)\}$:
\begin{equation}
\hat{A}_{i} = \frac{r(q,o_i) - \text{mean}(\{R_i\}_{i=1}^G)}{\text{std}(\{R_i\}_{i=1}^G)}.
\end{equation}
GRPO adopts a clipped objective and combines a KL penalty term:
\begin{equation}
\begin{aligned}
& \mathcal{J}_{\mathrm{GRPO}}(\theta)=\mathbb{E}_{(q, a) \sim \mathcal{D},\left\{o_i\right\}_{i=1}^G \sim \pi_{\theta \text { old }}(\cdot \mid q)} \\
& {\left[\frac { 1 } { G } \sum _ { i = 1 } ^ { G } \frac { 1 } { | o _ { i } | } \sum _ { l = 1 } ^ { | o _ { i } | } \left(\operatorname { m i n } \left(w_{i, l}(\theta) \hat{A}_{i},\right.\right.\right. \text { clip }} \\
& \left.\left.\left.\left(w_{i, l}(\theta), 1-\varepsilon, 1+\varepsilon\right) \hat{A}_{i}\right)-\beta D_{\mathrm{KL}}\left(\pi_\theta \| \pi_{\text {ref }}\right)\right)\right]
\end{aligned}
\end{equation}
where $w_{i, l}(\theta)=\frac{\pi_\theta\left(o_{i, l} \mid q, o_{i,<l}\right)}{\pi_{\theta_{\text {old }}}\left(o_{i, l} \mid q, o_{i,<l}\right)}$, $\pi_{\text{ref}}$ denotes the reference policy and $\beta$ controls the strength of the KL penalty. 

 % the token-level entropy at postion $k$ is defined as
\paragraph{Entropy Regularization.}
To alleviate the policy entropy collapse, a common approach is to explicitly incorporate entropy regularization as an auxiliary objective. For an autoregressive policy $\pi_\theta$,
the token-level entropy at position $l$ is defined as
\begin{equation}
\begin{aligned}
\mathcal H_l = -\!\!\sum_{o_l\in\mathcal V}\pi_\theta(o_l \mid q,o_{<l}) \,\log \pi_\theta(o_l \mid q,o_{<l}),
\end{aligned}
\end{equation}
where $\mathcal V$ denotes the vocabulary. The sequence-level entropy is the average over response positions:
\begin{equation}
\begin{aligned}
\mathcal H(\pi_\theta(\cdot \mid q)) = \frac{1}{L}\sum_{l=1}^L \mathcal H_l,
\end{aligned}
\end{equation}
with $L$ being the response length. The entropy regularization objective is given by
\begin{equation}
\begin{aligned}
\mathcal{J}_{\text{ent}}(\theta) = \gamma \,\mathbb{E}_{q\sim\mathcal D}\!\left[\mathcal H(\pi_\theta(\cdot \mid q))\right],
\end{aligned}
\end{equation}
where $\gamma$ controls the regularization strength. The overall objective combines this entropy regularization with the policy optimization objective:
\begin{equation}
\begin{aligned}
\mathcal{J}_{\text{total}}(\theta) = \mathcal{J}_{\text{GRPO}}(\theta) + \mathcal{J}_{\text{ent}}(\theta).
\end{aligned}
\end{equation}

\subsection{Adaptive Entropy Regularization (AER)}
\label{sec:aer}
AER dynamically balances exploration and exploitation in RLVR training by incorporating three complementary mechanisms:  
(i) \textbf{Difficulty-Aware Entropy Allocation}, which adjusts sample-level entropy coefficients according to task difficulty;  
(ii) \textbf{Initial-Anchored Target Entropy}, which adaptively defines the target exploration level based on the model’s initial entropy; and  
(iii) \textbf{Dynamic Global Coefficient Adjustment}, which continuously tunes the overall regularization strength to stabilize global policy entropy.  
These components form a self-regulating system that allocates exploration where needed, adapts to different setups, and maintains stable entropy dynamics. Formally, the training objective at iteration $t$ is:
\begin{equation}
\begin{aligned}
\label{eq:aer-total}
\mathcal{J}_{\text{AER}}(\theta)
&=\;\mathcal{J}_{\text{GRPO}}(\theta)\; \\
&+\;
\mathbb{E}_{q\sim\mathcal{D}}\!\big[\lambda_t(q)\,\mathcal{H}(\pi_\theta(\cdot\mid q))\big],
\end{aligned}
\end{equation}
where $\mathcal{H}(\pi_\theta(\cdot\mid q))$ denotes the sequence-level entropy and 
$\lambda_t(q)$ represents the adaptive coefficient for each sample.

\paragraph{C1. Difficulty-Aware Coefficient Allocation.}
As discussed in~\Secref{sec:pre_analysis}, harder questions require stronger exploration to uncover potential reasoning trajectories, whereas excessive exploration on easy questions can hinder convergence. To achieve this balance, we introduce a mechanism to estimate each question's difficulty relative to the current policy during training. GRPO naturally supports this estimation: for each question $q$, the policy generates a group of $m$ candidate responses $\{o_j\}_{j=1}^m$, 
from which we compute the \emph{group accuracy} as
\begin{equation}
\label{eq:group-acc}
g(q) = \frac{1}{m} \sum_{j=1}^{m} r(q, o_j), \quad r(q, o_j) \in \{0,1\}.
\end{equation}

% \vspace{-5mm}
\begin{table*}[ht]
\centering
\renewcommand{\arraystretch}{1.11}
\small
\begin{adjustbox}{width=\textwidth}
\begin{tabular}{
  l
  *{10}{S[table-format=2.1]}
}
\toprule
\multicolumn{1}{c}{\multirow[c]{2}{*}{\textbf{Method}}} 
  & \multicolumn{2}{c}{\textbf{AIME24}}
  & \multicolumn{2}{c}{\textbf{AIME25}}
  & \multicolumn{2}{c}{\textbf{AMC23}}
  & \multicolumn{2}{c}{\textbf{MATH500}}
  & \multicolumn{2}{c}{\textbf{Avg.}} \\
\cmidrule(lr){2-3}\cmidrule(lr){4-5}\cmidrule(lr){6-7}\cmidrule(lr){8-9}\cmidrule(lr){10-11}
& {\it pass@1} & {\it pass@32}
& {\it pass@1} & {\it pass@32}
& {\it pass@1} & {\it pass@32}
& {\it pass@1} & {\it pass@32}
& {\it pass@1} & {\it pass@32} \\
\midrule
\multicolumn{11}{c}{\textit{Qwen3-4B-Base}} \\
\midrule
Base            & 9.0  & 40.0 & 3.2  & 27.0 & 34.6 & 86.7 & 52.2 & 92.0 & 24.8 & 61.5 \\
\midrule
GRPO            & 19.7 & 42.0 & 15.4 & 32.3 & 59.9 & 87.8 & 80.5 & 93.7 & 43.9 & 64.0 \\
w/ Clip-Higher  & 20.4 & 32.0 & 20.0 & 40.3 & 59.6 & 89.7 & 81.1 & 93.5 & 45.3 & 63.9 \\
w/ Ent-Adv      & 21.0 & 40.9 & 13.3 & 31.6 & 49.2 & 86.7 & 79.2 & 91.3 & 40.7 & 62.6 \\
w/ KL-Cov       & 24.5 & \bfseries 54.6 & 20.4 & 39.2 & 57.2 & 84.9 & 80.3 & 95.5 & 45.6 & 68.6 \\
w/ Clip-Cov     & 24.0 & 52.5 & \bfseries 22.5 & 41.4 & 65.9 & 92.4 & \bfseries 87.8 & 95.9 & 50.1 & 70.6 \\
\rowcolor{gray!18}
w/ AER (Ours)   & \bfseries 25.2 & 49.6 & 22.1 & \bfseries 48.9 & \bfseries 70.2 & \bfseries 94.8 & 86.7 & \bfseries 96.6 & \bfseries 51.1 & \bfseries 72.5 \\
\midrule
\multicolumn{11}{c}{\textit{Qwen3-8B-Base}} \\
\midrule
Base            & 11.5 & 48.0 & 8.8  & 35.5 & 45.0 & 88.7 & 67.2 & 94.2 & 33.1 & 66.6 \\
\midrule
GRPO            & 20.5 & 47.0 & 18.5 & 34.3 & 62.8 & 88.4 & 82.0 & 94.1 & 46.0 & 66.0 \\
w/ Clip-Higher  & 27.1 & 61.5 & 21.1 & 46.9 & 68.5 & 92.7 & 88.3 & 97.0 & 51.3 & 74.5 \\
w/ Ent-Adv      & 23.2 & 45.0 & 15.8 & 35.2 & 62.5 & 80.6 & 83.5 & 93.4 & 46.3 & 63.6 \\
w/ KL-Cov       & 26.9 & 58.4 & 20.8 & 42.7 & 65.3 & 92.1 & 80.6 & 96.7 & 48.4 & 72.5 \\
w/ Clip-Cov     & 30.8 & 62.1 & 24.4 & 46.9 & 74.5 & 94.1 & \bfseries 90.4 & \bfseries 97.8 & 55.0 & 75.2 \\
\rowcolor{gray!18}
w/ AER (Ours)   & \bfseries 31.4 & \bfseries 63.2 & \bfseries 25.1 & \bfseries 48.9 & \bfseries 75.6 & \bfseries 94.8 & 89.4 & 97.0 & \bfseries 55.4 & \bfseries 76.0 \\
\bottomrule
\end{tabular}
\end{adjustbox}
\caption{\textbf{Overall performance comparison.} We compared the \textit{pass@1} and \textit{pass@32} performance between different methods using the Qwen3-4B-Base and Qwen3-8B-Base models on four mathematical reasoning benchmarks.}
\label{tab:main_res}
\end{table*}

\textbf{Allocation rule.}
Given a pivot accuracy $\rho \in[0,1]$ and a small constant $\varepsilon>0$ (typically $10^{-8}$), 
we define a \emph{difficulty-aware entropy coefficient} as
\begin{equation}
\begin{aligned}
\label{eq:lambda-onesided}
\lambda_t(q)& = 
\alpha_t \cdot \frac{\max\{0,\,\rho - g(q)\}}{\rho+\varepsilon} \\
\;&+\;
\alpha_t \cdot \mathbb{I}\{\rho=0,\,g(q)=0\},
\end{aligned}
\end{equation}
where $\alpha_t$ is a \emph{global entropy scaling factor} controlling the overall strength of entropy regularization, which is adaptively updated through the dynamic adjustment mechanism introduced in C3.  
The coefficient $\lambda_t(q)$ is positive only for hard questions ($g(q)\le\rho$) and zero otherwise. Additionally, it allocates larger entropy coefficients to more difficult questions based on the difference between the current $ g(q) $ and $ \rho $ to encourage more exploration.

\paragraph{C2. Initial-Anchored Target Entropy.}
As analyzed in~\Secref{sec:pre_analysis}, maintaining the policy entropy within a certain ``sweet spot'' below its initial value can appropriately promote exploration and stabilize training.
However, the initial policy entropy varies substantially across base models, datasets, and sampling temperature, making it difficult to directly specify a specific value as the target entropy. Therefore, we seek an adaptive mechanism that can determine the value of the target entropy based on the initial entropy at each run.

\textbf{Definition.}
Let $\Hglob_0$ denote the policy entropy at initialization (obtained in the first training step).  
We define the \emph{initial-anchored target entropy} as
\begin{equation}
\label{eq:target-entropy}
H^\star = \tau \cdot \Hglob_0, \qquad \tau \in (0,1),
\end{equation}
where $\tau$ is a predefined reduction ratio representing the desired relative decrease from the initial entropy. 
By anchoring the target entropy to the initial entropy, it adaptively calibrates the effective ``exploration budget'' across different setups, thereby alleviating the burden of repeated hyperparameter tuning and improving cross-run stability.

\paragraph{C3. Dynamic Global Coefficient Adjustment.}\label{sec:C3}
While C1 determines \emph{where} to allocate exploration and C2 specifies \emph{how much} entropy should be retained, 
the global policy entropy can still drift from the target $H^\star$ as training proceeds. 
Recall that $\alpha_t$ in~\Eqref{eq:lambda-onesided} serves as a global scaling factor that controls the overall strength of entropy regularization. 
Instead of keeping $\alpha_t$ fixed as a hyperparameter, AER updates it dynamically according to the observed policy entropy~\citep{he2025skywork}. 
This mechanism follows the classical principle of \emph{closed-loop control}~\citep{aastrom2021feedback}, 
in which a system continuously measures its output, compares it to a target, and adjusts its input to reduce the deviation. 

\textbf{Definition.}
Let $\Hglob_t$ denote the policy entropy at iteration $t$.  
With step size $\eta>0$, the global entropy scaling factor $\alpha_t$ in~\Eqref{eq:lambda-onesided} is updated as
\begin{equation}
\label{eq:alpha-update}
\alpha_{t+1} = \big[\alpha_t + \eta \cdot \mathrm{sgn}(H^\star - \Hglob_t)\big]_+,
\end{equation}
where $[z]_+=\max(z,0)$. If $\Hglob_t < H^\star$, $\alpha_t$ increases to encourage exploration; if $\Hglob_t > H^\star$, $\alpha_t$ decreases to suppress excessive exploration.  
% This behavior resembles a thermostat that adjusts its output to maintain temperature near the desired setpoint— an intuitive analogy for how AER automatically regulates the ``entropy temperature'' of the policy.

This dynamic global coefficient adjustment prevents both \emph{entropy collapse} and \emph{entropy explosion}. By continuously monitoring and correcting deviations, AER maintains policy entropy near $H^\star$, forming a dynamic self-regulating mechanism that sustains balanced exploration without much manual tuning of hyperparameters.

\paragraph{Workflow of AER.}
Each training iteration of AER proceeds as follows:
(i) estimate the group accuracy $g(q)$ for each question via~\Eqref{eq:group-acc};
(ii) compute difficulty-aware coefficients $\lambda_t(q)$ via ~\Eqref{eq:lambda-onesided};
(iii) optimize $\mathcal{J}_{\text{AER}}(\theta)$ in ~\Eqref{eq:aer-total}; and 
(iv) update $\alpha_t$ according to ~\Eqref{eq:alpha-update}.  
This closed-loop process adaptively allocates and regulates exploration, maintaining stable entropy dynamics throughout RLVR training.

\begin{figure*}[t]
  \centering
  % ---- Row 1 ----
  \begin{subfigure}[t]{0.31\textwidth}
    \centering
    \includegraphics[width=\linewidth]{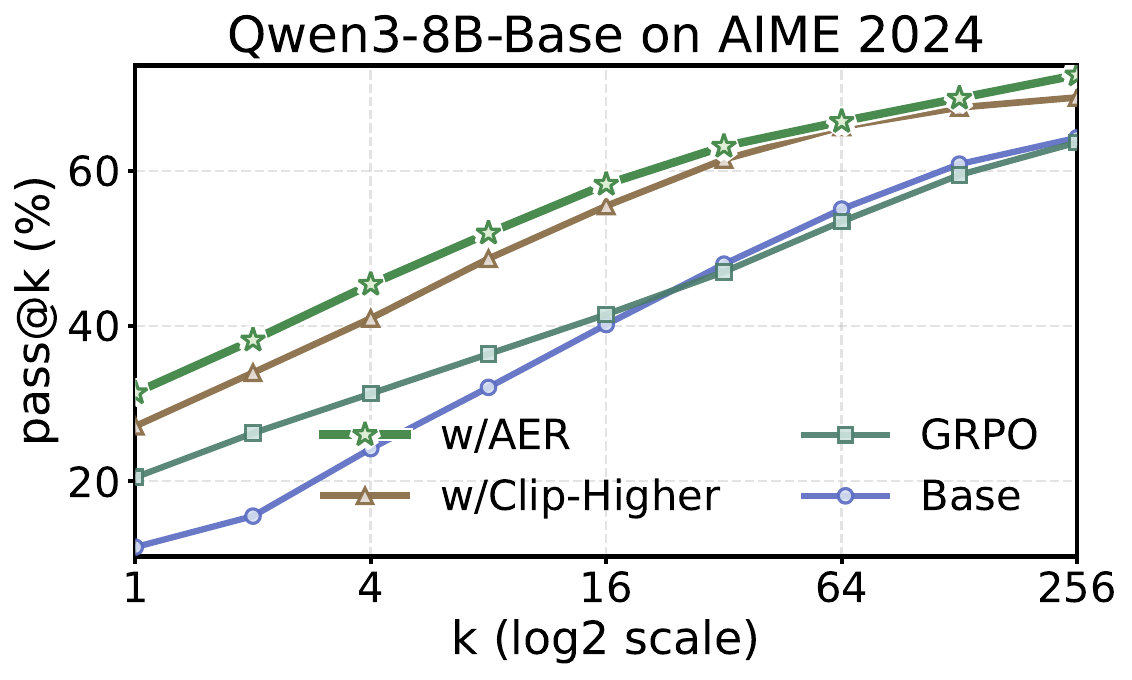}
    \caption{Pass@k on AIME 2024}
    \label{fig:stack_a}
  \end{subfigure}\hfill
  \begin{subfigure}[t]{0.31\textwidth}
    \centering
    \includegraphics[width=\linewidth]{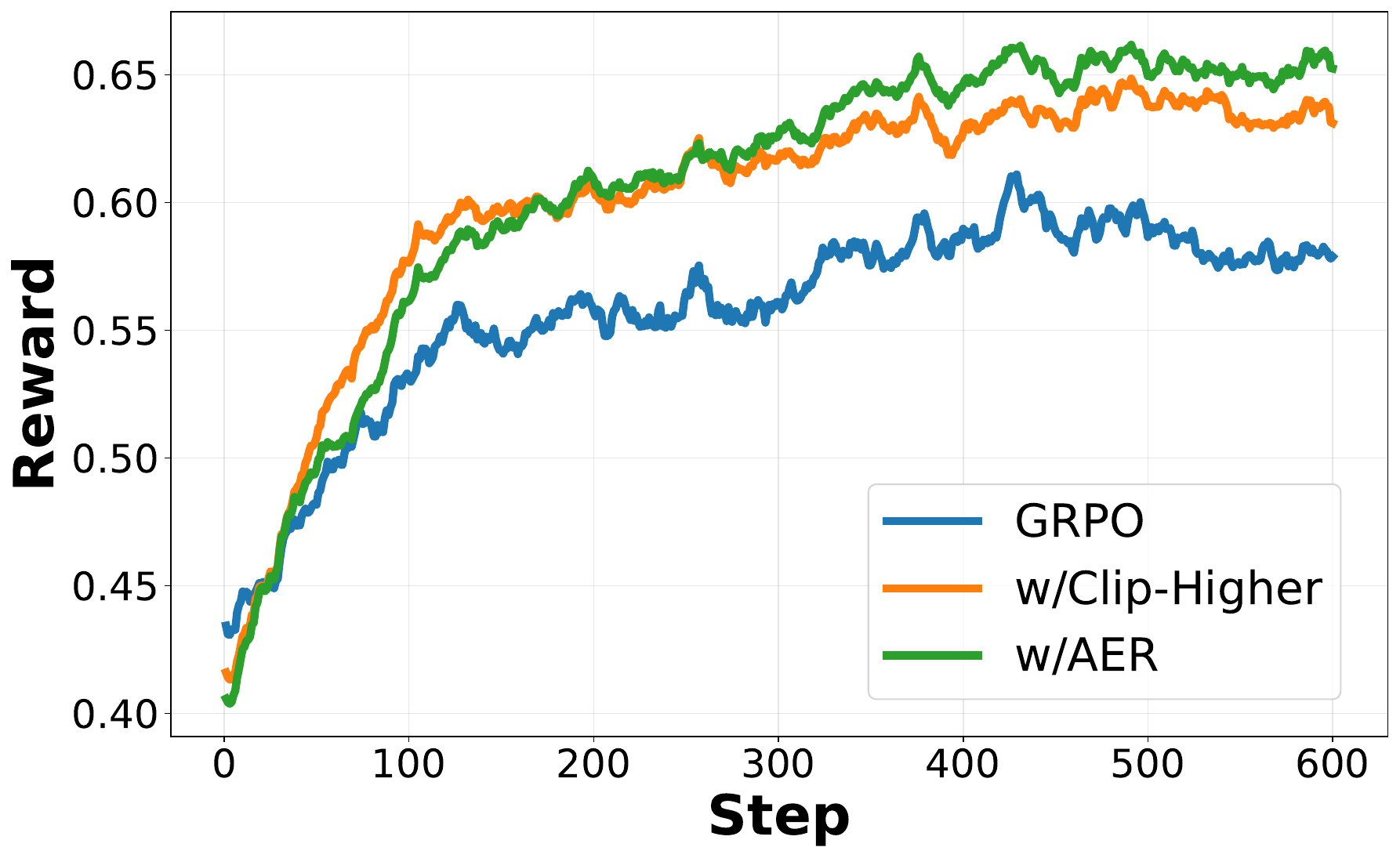}
    \caption{Training Reward}
    \label{fig:stack_b}
  \end{subfigure}\hfill
  \begin{subfigure}[t]{0.31\textwidth}
    \centering
    \includegraphics[width=\linewidth]{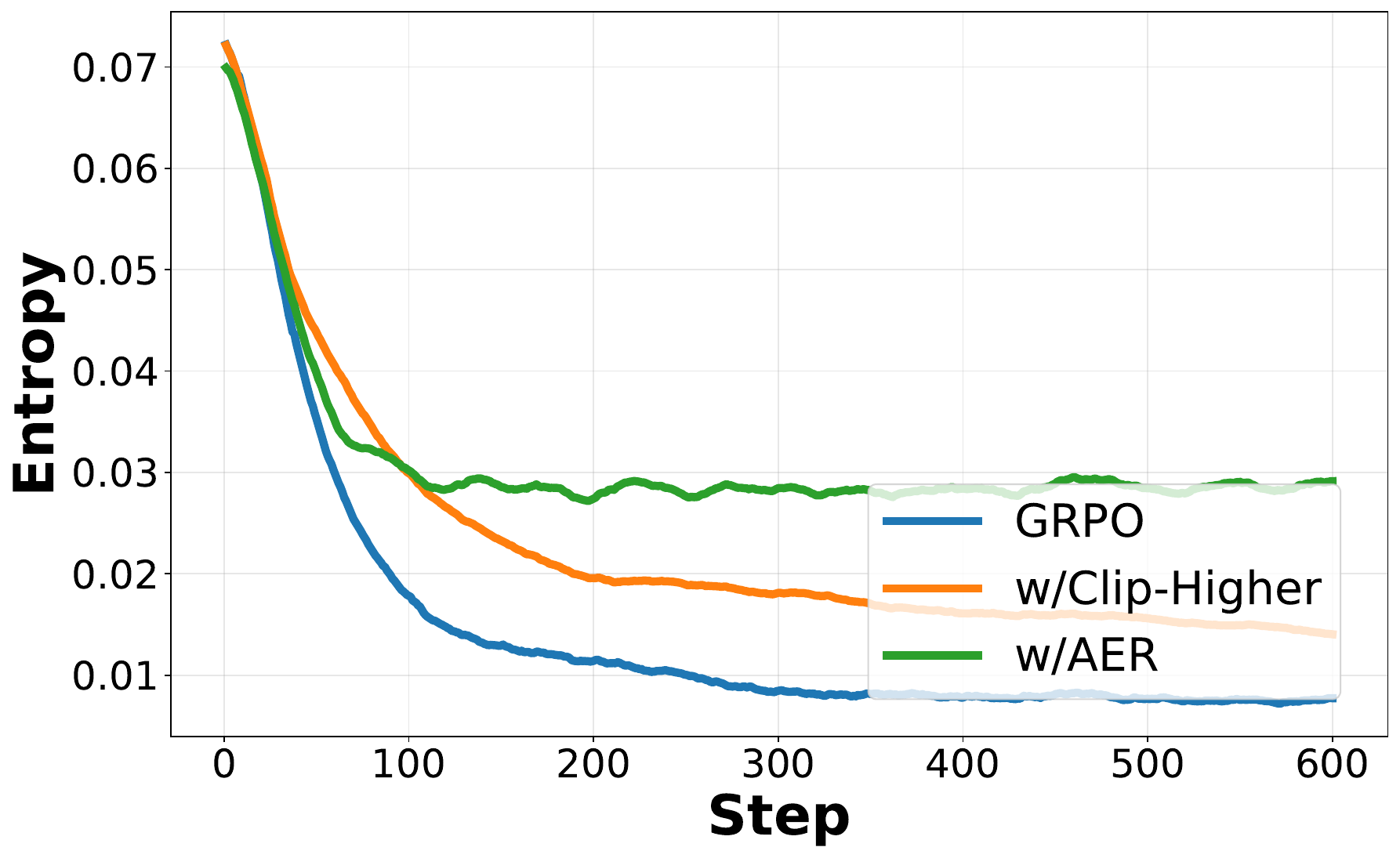}
    \caption{Policy Entropy}
    \label{fig:stack_c}
  \end{subfigure}

  \vspace{0.3em}

  % ---- Row 2 ----
  \begin{subfigure}[t]{0.31\textwidth}
    \centering
    \includegraphics[width=\linewidth]{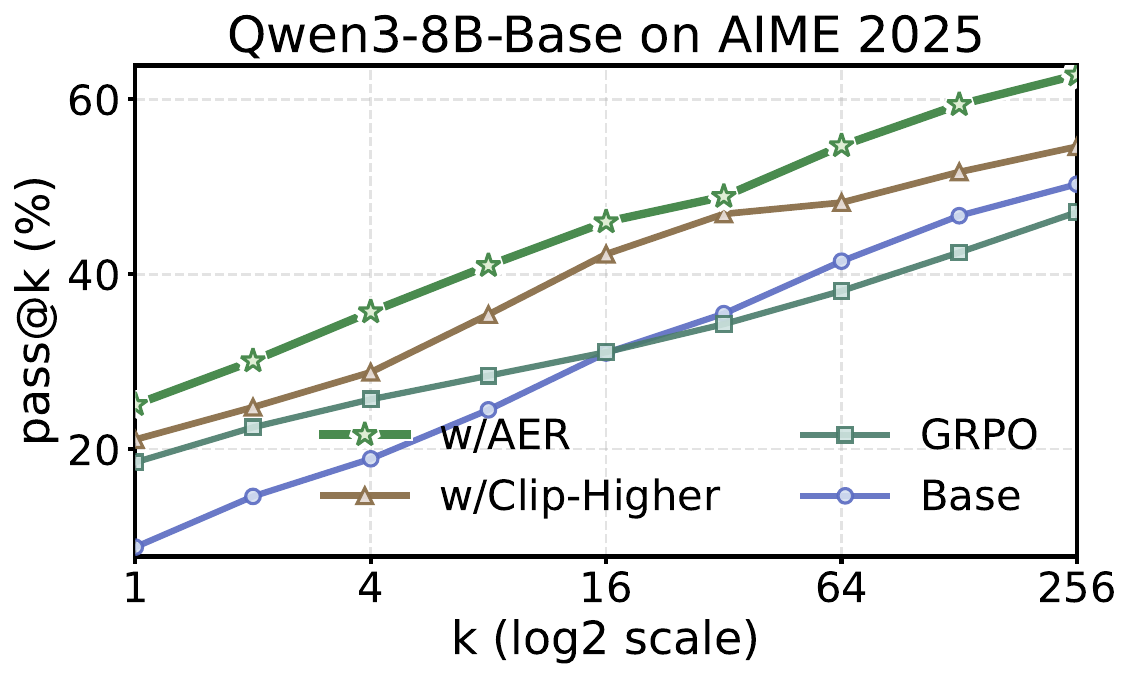}
    \caption{Pass@k on AIME 2025}
    \label{fig:stack_d}
  \end{subfigure}\hfill
  \begin{subfigure}[t]{0.31\textwidth}
    \centering
    \includegraphics[width=\linewidth]{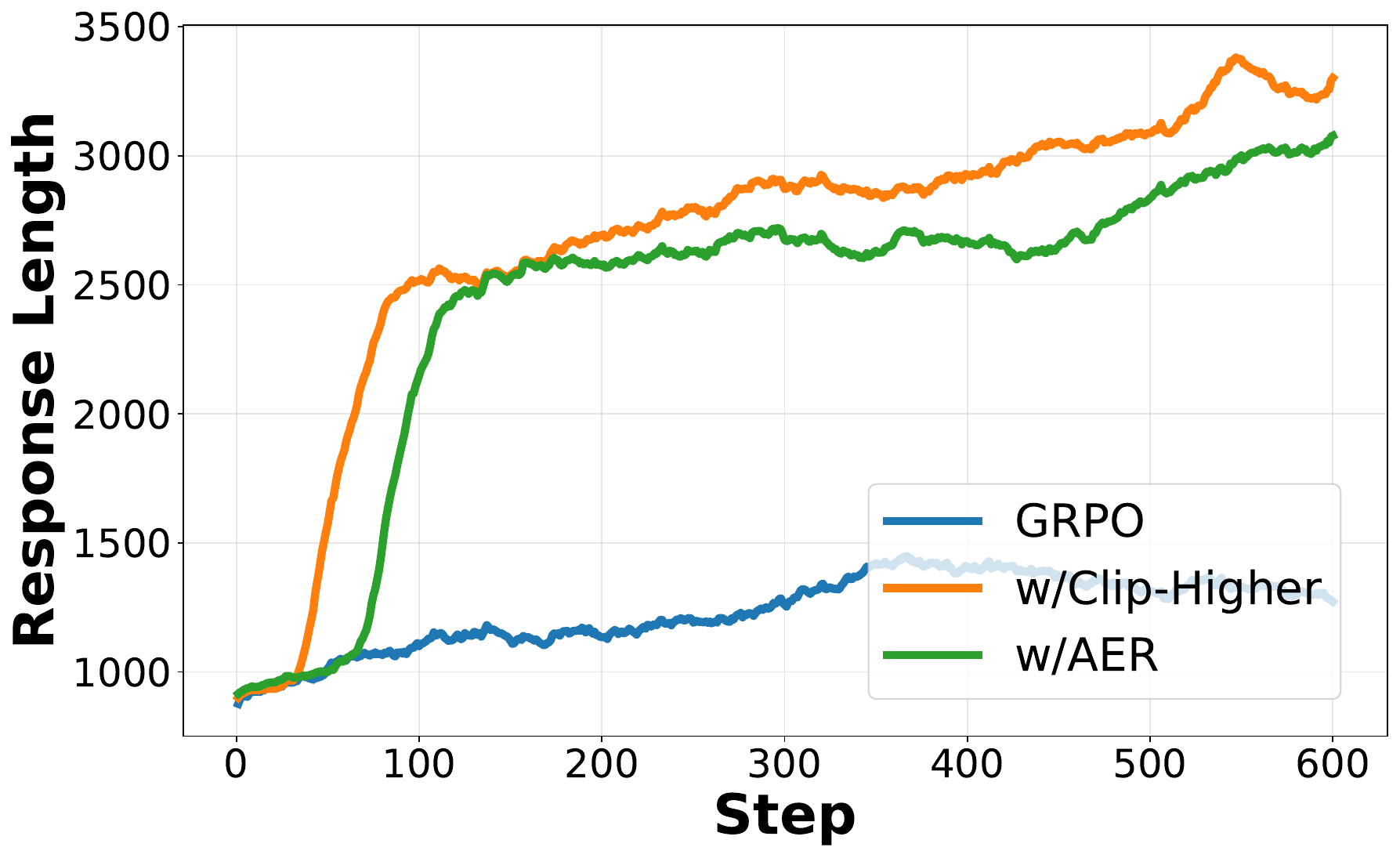}
    \caption{Avg. Response Length}
    \label{fig:stack_e}
  \end{subfigure}\hfill
  \begin{subfigure}[t]{0.31\textwidth}
    \centering
    \includegraphics[width=\linewidth]{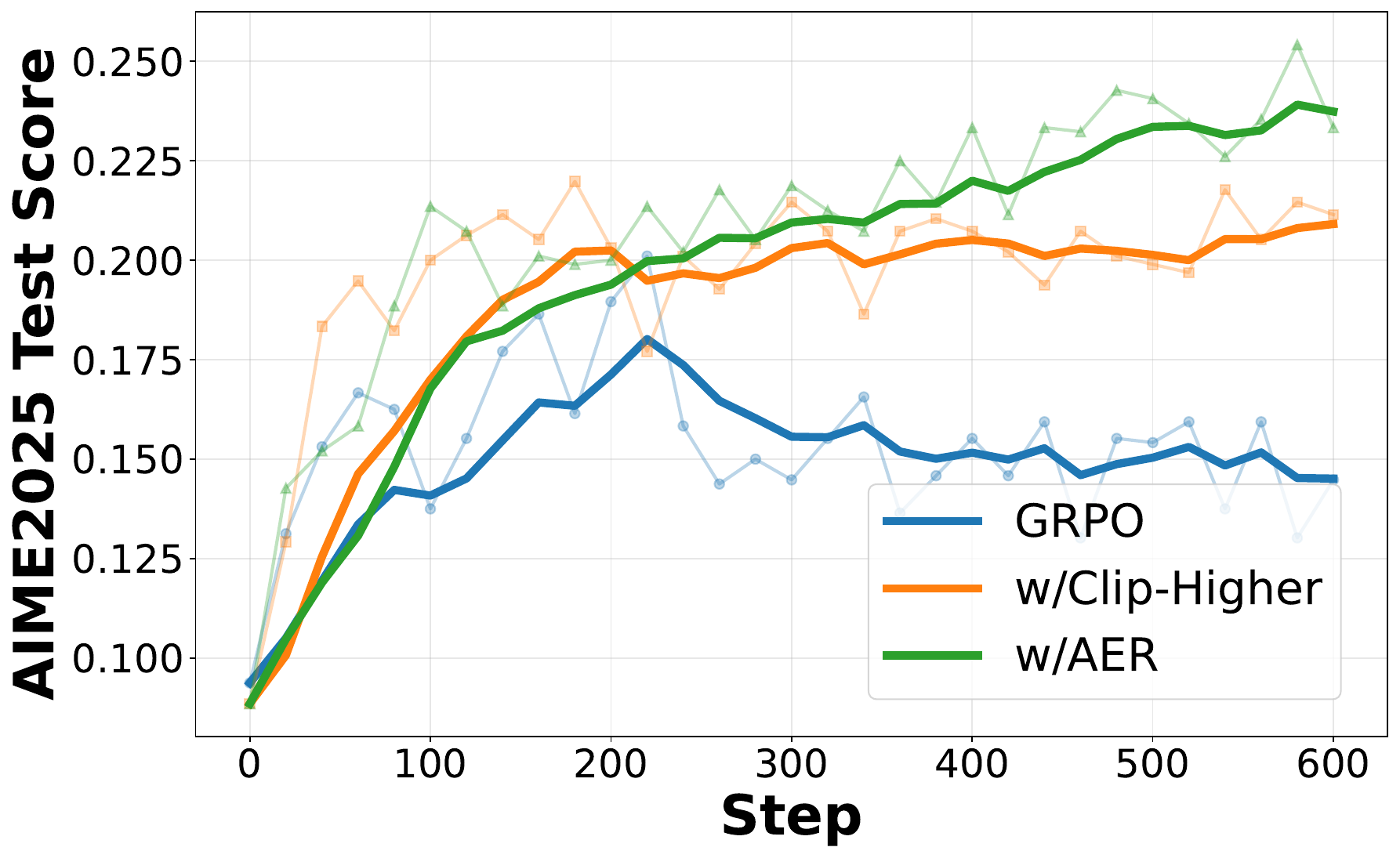}
    \caption{Test Score on AIME25}
    \label{fig:stack_f}
  \end{subfigure}

  \caption{\textbf{Pass@$k$ and training dynamics (2×3 grid).}
  Left column: pass@$k$ on AIME24 (a) and AIME25 (d) as $k$ scales.
  Right two columns: training dynamics—reward (b), policy entropy (c), response length (e)—and test score on AIME25 (f) over steps.
  }
  \label{fig:overview_2x3}
\end{figure*}

\section{Experiments}
\label{tab:experiments}

\subsection{Experiment Setup}
\paragraph{Configurations.} 
We adopt the widely used Qwen series open-source models~\cite{yang2025qwen3} for validating the efficacy of the proposed method, including Qwen3-4B-Base and Qwen3-8B-Base. All models are trained on the open-source DeepscaleR dataset~\citep{luo2025deepscaler}. 
We use the Hugging Face verification tool \texttt{math\_verify}~\citep{mathverify2025} to automatically check the correctness of model answers.
Detailed hyperparameter configuration  can be found in Appendix~\ref{app:trainingsetup}.

\paragraph{Evaluations.} 
We follow standard protocols to evaluate mathematical reasoning and select widely adopted benchmarks: AIME24~\citep{aime24}, AIME25~\citep{aime25},  AMC23~\citep{amc23} and  MATH500~\citep{hendrycks2021measuring}. We report pass@1 to evaluate performance and pass@$k$ to evaluate exploration capability~\citep{yue2025does}. Greater exploration improves the likelihood of finding a correct reasoning path within $k$ attempts.

\paragraph{Baseline methods.}
We compare AER against several baselines, including the vanilla GRPO~\citep{shao2024deepseekmath}, GRPO with Clip-Higher~\citep{yu2025dapo} and several advanced exploration-oriented methods: Ent-Adv~\citep{cheng2025reasoning}, which introduces an entropy-based advantage term to encourage longer reasoning trajectories, and Clip-Cov and KL-Cov~\citep{cui2025entropy}, which regularize policy updates through token–entropy covariance to stabilize training.

\subsection{Main Results}
\paragraph{Overall Performance.}
Table~\ref{tab:main_res} presents the main results on four mathematical reasoning benchmarks.
Across both model scales, \textbf{AER} generally achieves higher performance (\textit{pass@1}) and stronger exploration capability (\textit{pass@32}) than the baselines. For the Qwen3-4B-Base model, AER yields an average \textbf{+7.2\%} improvement in \textit{pass@1} over vanilla GRPO and \textbf{+1.0\%} over the best baseline (Clip-Cov). In terms of exploration, AER improves the average \textit{pass@32} by \textbf{+8.5\%} compared to vanilla GRPO and \textbf{+1.9\%} compared to Clip-Cov.
A similar trend is observed for the Qwen3-8B-Base model, where AER reaches the highest average scores—\textbf{55.4\%} on \textit{pass@1} and \textbf{76.0\%} on \textit{pass@32}. The gains are more noticeable on challenging benchmarks such as AIME24 and AIME25, suggesting that difficulty-aware entropy allocation may better promote exploration on harder reasoning tasks.
Overall, the results indicate that AER can enhance reasoning performance and exploration diversity, highlighting its potential to further leverage entropy regularization in RLVR training for LLMs. For additional experimental results, including scalability analysis on larger models and generalization to other architectures, please refer to Appendix~\ref{app:detailed-results}.

% pass_at_k
% \subsection{Pass@$k$ Analysis.}
\begin{figure*}[t]
  \centering
  \begin{subfigure}[t]{0.24\textwidth}
    \centering
    \includegraphics[width=\linewidth]{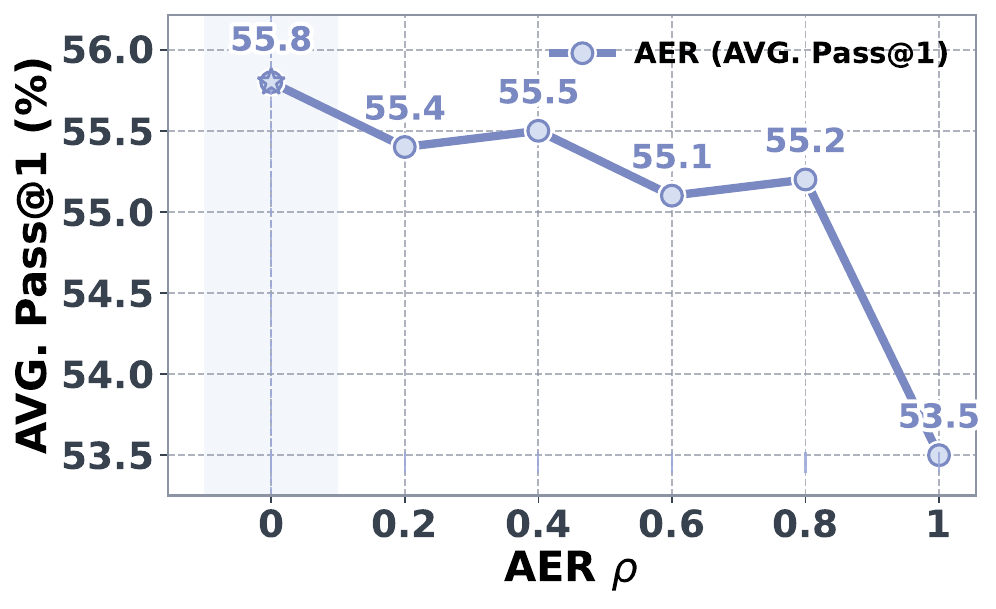}
    \caption{AER $\rho$ ablation}
    \label{fig:exp_aa}
  \end{subfigure}\hfill
  \begin{subfigure}[t]{0.24\textwidth}
    \centering
    \includegraphics[width=\linewidth]{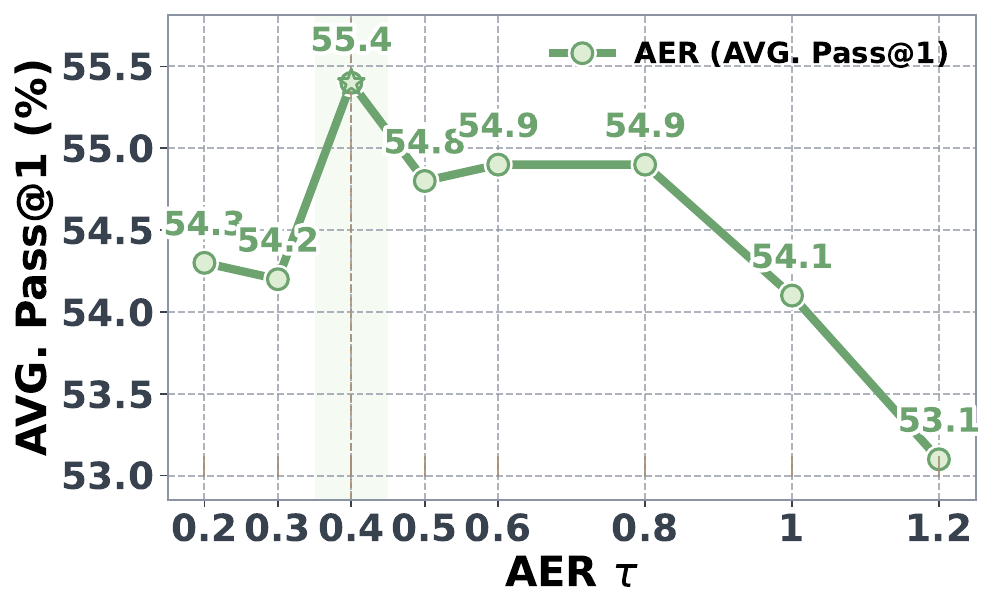}
    \caption{AER $\tau$ ablation}
    \label{fig:exp_ab}
  \end{subfigure}\hfill
  \begin{subfigure}[t]{0.24\textwidth}
    \centering
    \includegraphics[width=\linewidth]{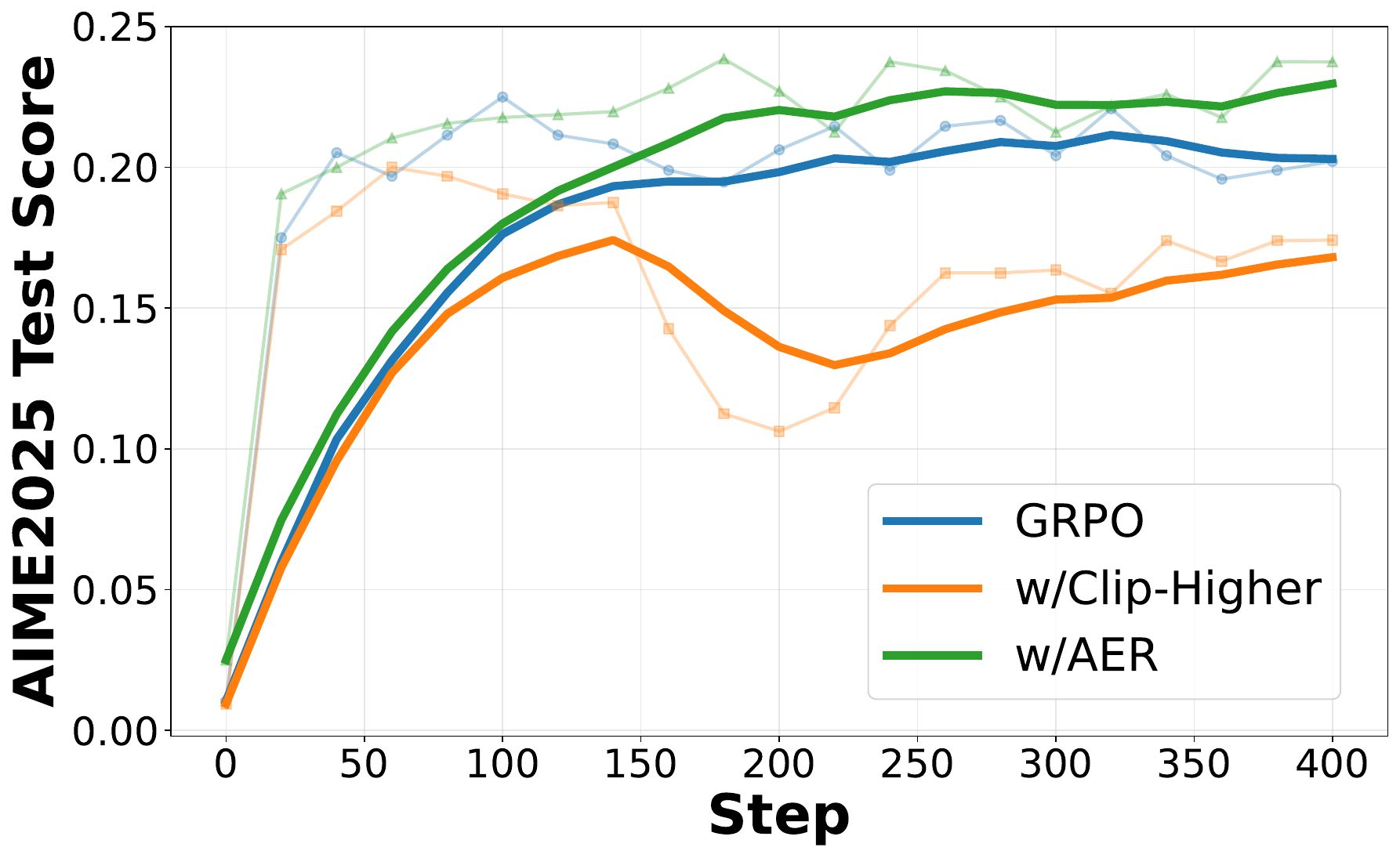}
    \caption{Qwen3-8B Test Score}
    \label{fig:exp_ac}
  \end{subfigure}\hfill
  \begin{subfigure}[t]{0.24\textwidth}
    \centering
    \includegraphics[width=\linewidth]{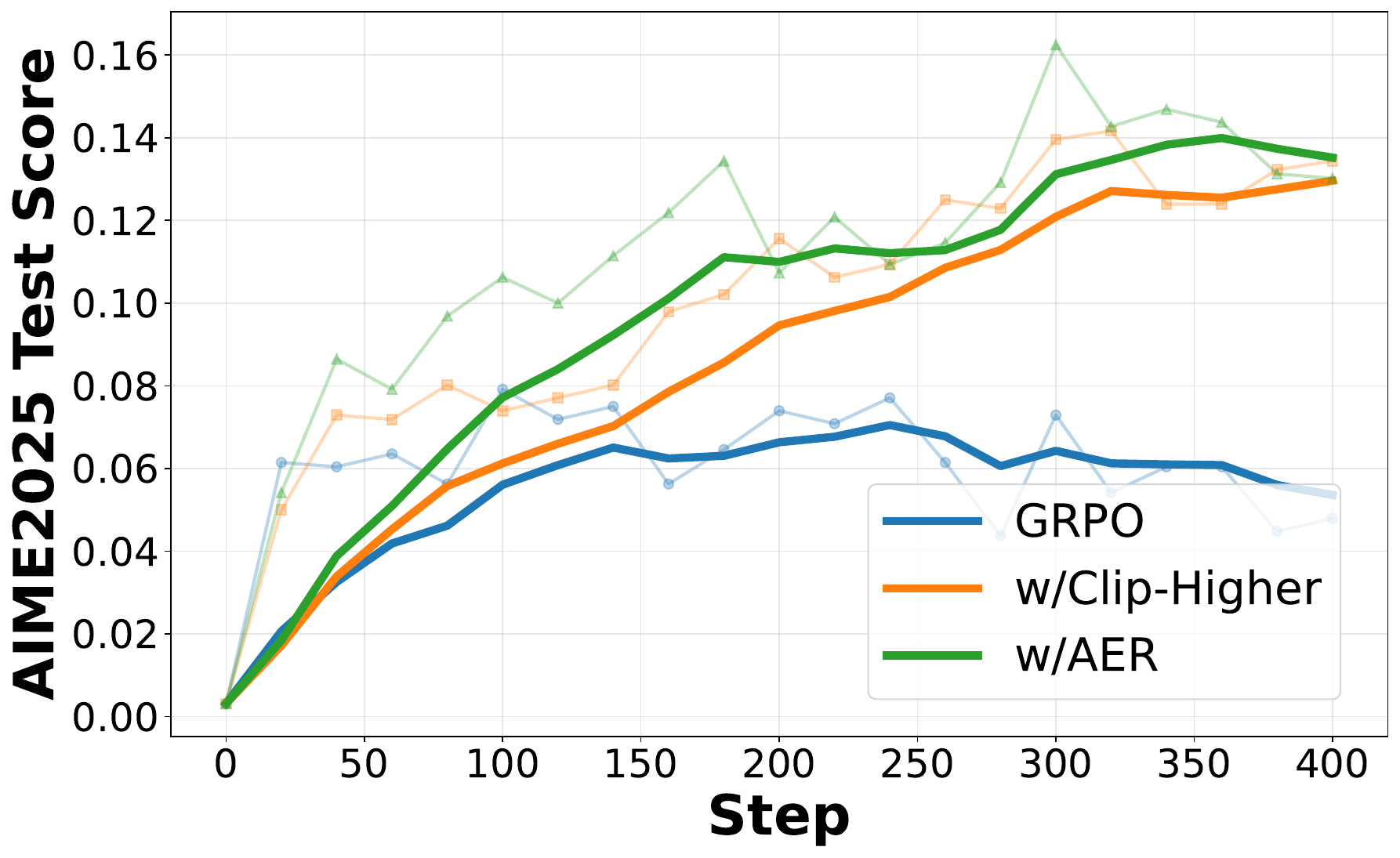}
    \caption{Qwen2.5-7B Test Score}
    \label{fig:exp_ad}
  \end{subfigure}
  % \caption{\textbf{Placeholder.} We xxxxxxxxxxxx}
  \caption{\textbf{AER ablations and generalization.}
(a) Pivot accuracy $\rho$ ablation (C1, $\tau{=}0.4$).
(b) Reduction ratio $\tau$ ablation (C2, $\rho{=}0.2$).
(c) Generalization on \textsc{dapo\_math\_17k}: Qwen3-8B-Base test score.
(d) Generalization on \textsc{dapo\_math\_17k}: Qwen2.5-7B-Base test score.}
  \label{fig:exp_a}
\end{figure*}

% 依然自适应 ACL 单栏宽度；去掉 Avg. 一列
\begin{table}[]
\centering
\LARGE
\renewcommand{\arraystretch}{1.1}
\setlength{\tabcolsep}{2.5pt}
\resizebox{1.0\linewidth}{!}{
\begin{adjustbox}{width=\columnwidth}
\begin{tabular}{l *{4}{S[table-format=2.1]}}
\toprule
\multicolumn{1}{c}{\textbf{Qwen3-8B-Base}} &
\multicolumn{1}{c}{\textbf{AIME24}} &
\multicolumn{1}{c}{\textbf{AIME25}} &
\multicolumn{1}{c}{\textbf{AMC23}} &
\multicolumn{1}{c}{\textbf{MATH500}} \\
\midrule
GRPO                       & 20.5 & 18.5 & 62.8 & 82.0 \\
w/ Fixed Coeff.             & 23.4 & 21.9 & 63.9 & 86.1 \\
w/ Difficulty-Aware. (C1) & 28.9 & 23.5 & 73.6 & 88.9 \\
w/ Adaptive Coeff. (C2\&C3) & 29.4 & 21.4 & 73.8 & 89.4 \\
w/ AER (C1\&C2\&C3)         & \bfseries 31.4 & \bfseries 25.1 & \bfseries 75.6 & \bfseries 89.4 \\
\bottomrule
\end{tabular}
\end{adjustbox}
}
\caption{Ablation study of different components.}
\label{tab:component_ablation}
\end{table}

\paragraph{Pass@$k$ Analysis.}
Following prior work~\citep{yue2025does, cheng2025reasoning, jiang2025rethinking}, we further examine the exploration boundary of different methods through pass@$k$ performance, which reflects the model's upper bound of reasoning capability.
Specifically, we evaluate the Qwen3-8B-Base model on two challenging benchmarks—AIME2024 and AIME2025—extending $k$ up to $256$ to assess how performance scales with increased rollout attempts. As shown in~\Figref{fig:stack_a} and~\Figref{fig:stack_d}, AER consistently achieves higher pass@$k$ performance across all $k$, indicating stronger exploration and reasoning diversity.
The advantage becomes more pronounced as $k$ increases on the AIME25 dataset, where AER maintains a clear margin over baselines throughout the curve.
This trend suggests that AER more effectively broadens the search space of reasoning trajectories, enabling the policy to discover a larger proportion of correct solutions under extended sampling. Detailed results are shown in Appendix~\ref{app:pass_at_k}.

\paragraph{Analysis of Training Dynamics.}
In~\Figref{fig:stack_b}, the training reward of AER reaches a higher level than the baselines, suggesting that our method facilitates more effective policy improvement during training.
As shown in~\Figref{fig:stack_c}, the policy entropy of vanilla GRPO drops sharply within the first 100 steps, indicating early entropy collapse. Although the baseline slightly mitigates this effect, its entropy still gradually declines over steps. AER maintains the policy entropy at a moderate target level throughout training, which help sustain consistent and balanced exploration. Despite maintaining higher entropy, AER produces shorter average response lengths than the baseline as shown in~\Figref{fig:stack_e}, implying that our method guides exploration more effectively without unnecessary verbosity. Finally,~\Figref{fig:stack_f} shows that AER dynamically maintains a stable policy entropy throughout training and allocates more exploration to difficult samples, thereby facilitating the policy to escape local optima and continuously improve performance. Overall, these results suggest that AER promotes stable entropy dynamics, balanced exploration, and sustained performance gains during RLVR training.

\subsection{Ablation Study}
\paragraph{Effectiveness of different components.}
As shown in \Tabref{tab:component_ablation}, building upon GRPO, introducing entropy regularization with a tuned fixed coefficient improves performance, suggesting that moderate exploration is beneficial. Then applying C2 \& C3 brings further gains, particularly on challenging benchmarks such as AIME24, highlighting the importance of maintaining an appropriate level of policy entropy throughout training.
When the C1 is further incorporated, the full AER framework achieves the best performance, e.g., improving AIME25 performance from $21.4$ to $25.1$.

\paragraph{Ablation of the pivot accuracy ($\rho$) in C1.}
Fixing $\tau{=}0.4$, we sweep $\rho$ to examine how it allocates exploration (\Figref{fig:exp_aa}). The \emph{Avg.\ pass@1} curve increases as $\rho$ decreases from $1.0$ to $0.0$, peaking at \textbf{55.8} when $\rho{=}0$, and degrading once $\rho\!\ge\!0.6$ (down to \textbf{53.5} at $\rho{=}1.0$). The monotonic $1\!\to\!0$ rise confirms our \emph{difficulty-aware} design: lowering $\rho$ gates positives to truly hard cases, focusing entropy where exploration helps while sparing easy ones. Although C3 stabilizes the global entropy around the target, an overly large $\rho$ makes many questions ``hard,” diluting per-sample selectivity and weakening the signal-to-noise ratio of entropy injection. Accordingly, we recommend a \textbf{small–moderate} pivot and adopt $\rho{<=}0.2$ by default.

\paragraph{Ablation of the reduction ratio ($\tau$) in C2.}

Fixing $\rho{=}0.2$, we sweep $\tau$ to study how anchoring the target entropy to the initial entropy affects performance (\Figref{fig:exp_ab}). The \emph{Avg.\ pass@1} curve exhibits a clear sweet spot at $\tau{=}0.4$ (\textbf{55.4}), with a mild plateau for $\tau\!\in\![0.5,0.6]$ and degradation in both directions: \emph{under}-reduction at $\tau\!\le\!0.3$ and \emph{over}-preservation at high-entropy settings $\tau\!\ge\!0.8$. These results substantiate the empirical premise that \emph{moderate} entropy reduction accompanies accuracy gains: compared with vanilla GRPO, AER stabilizes the global entropy \emph{near} the prescribed band, retaining sufficient diversity while avoiding premature collapse. Pushing $\tau$ too high inflates entropy beyond the useful range, diluting the signal for exploitation; pushing it too low curtails beneficial exploration. In practice, we thus adopt $\tau\!\in\![0.4,0.6]$ as default.

\paragraph{Generalization under different experimental settings.}
To test whether AER’s closed-loop entropy control transfers across data and architectures without retuning, we replace \textsc{DeepscaleR} with \textsc{dapo\_math\_17k} and train \textbf{Qwen3-8B-Base} and \textbf{Qwen2.5-7B-Base} using the same default hyperparameters (\(\rho{=}0.2,\ \tau{=}0.4\)). The test-score curves in \Figref{fig:exp_ac} and \Figref{fig:exp_ad} show that AER achieves faster early gains and maintains a clear, persistent margin over \textit{Clip-Higher} throughout training, while converging to a higher (or comparable) asymptote than \textit{GRPO}. This behavior indicates that the initial-anchored target entropy (C2) together with dynamic global scaling (C3) adapts the effective exploration level to the new data distribution and model backbone.

\section{Conclusion}
This paper revisits entropy regularization in reinforcement learning with verifiable rewards and identifies fixed-coefficient schemes—often prone to entropy collapse or explosion—as a key limitation to reasoning performance. We introduce \textit{Adaptive Entropy Regularization (AER)}, which adaptively balances exploration and exploitation throughout training. On multiple mathematical reasoning benchmarks, AER consistently improves both performance and exploration capability, revealing that adaptive coefficient control is crucial to rediscovering the potential of entropy regularization for LLM reinforcement learning.

\section*{Limitations}
\label{sec:limitations}
We discuss one practical limitation of our work. 
In line with most open-source studies, we train primarily on mathematical datasets because they are easy to obtain and provide accurate, verifiable rewards. This is not an important issue, and we believe they can achieve better results if more training datasets with verifiable rewards for general domains are available.

\section*{Ethics Statement}

We have carefully considered the ethical implications of our research and provide the following statements:(1) Throughout this study, we have strictly followed established ethical guidelines, ensuring that our findings are reported honestly, transparently, and with full accuracy.
(2) No sensitive or confidential information was used at any stage of our research. All data and materials utilized are suitable for public release.
(3) The datasets employed in our experiments originate from publicly available and peer-reviewed scientific sources, supporting the transparency and reproducibility of our work.
(4) We offer detailed descriptions of the datasets and the hyper-parameter configurations used in our experiments to ensure the reproducibility and clarity of our results.
(5) In the interest of openness and to support future research, we have made our code available anonymously on GitHub and will fully open source it following the acceptance of our paper.

\bibliography{ref_bib}

\clearpage
\appendix

% \onecolumn

\section{Implementation Details}
\label{app:trainingsetup}

% \clearpage
\subsection{Training Setup}
We conduct all experiments on a single node equipped with \textbf{32$\times$ NVIDIA H100 (80\,GB, SXM)} GPUs interconnected via NVLink. Unless otherwise noted, all methods are trained under the same computational budget (wall-clock hours and effective batch size) and with identical data pre-processing and evaluation protocols to ensure fair comparison.

\paragraph{RL stack and integration.}
For the reinforcement learning (RL) stage, we adopt \textsc{VeRL}, a post-training framework tailored to large language models (LLMs) with verifiable or preference-based feedback. \textsc{VeRL} provides modular APIs that integrate seamlessly with mainstream LLM infrastructures (e.g., PyTorch FSDP and Megatron-LM) and inference engines (e.g., vLLM), enabling (i) memory-efficient sharding for optimizer states and activations, (ii) flexible rollout orchestration, and (iii) pluggable algorithm components (e.g., GRPO, DAPO, and our variants). This design reduces engineering overhead and makes experimental factors (algorithmic choices, hyperparameters, and decoding policies) \emph{isolatable and reproducible}.

\paragraph{Distributed training and precision.}
We use mixed-precision training (bf16 where supported, with automatic casting and safe gradient scaling) and distributed data parallelism via FSDP-style full sharding of parameters, gradients, and optimizer states. Gradient accumulation is employed to match the effective global batch sizes reported in \autoref{tab:train_config}. Checkpointing is performed at fixed intervals to support fault tolerance and ablations with matched training budgets.

\paragraph{Inference engine for rollouts and evaluation.}
All online rollouts and offline evaluations are executed with \textsc{vLLM}, an efficient LLM inference engine that supports asynchronous batching and distributed serving with a paged key–value (KV) cache. Using a single, shared engine for both training-time rollouts and test-time evaluation minimizes distribution shift induced by heterogeneous runtimes. Decoding configurations (temperature, nucleus/top-$p$, maximum length, stop rules) are held \emph{constant across methods} within each experiment; pass@$k$ metrics use a fixed $k$ (default $k{=}32$) unless otherwise specified.

\paragraph{Reproducibility.}
We fix random seeds for data sampling, parameter initialization, and decoder sampling; we further report the exact learning rate, context length, evaluation interval, and method-specific knobs in \autoref{tab:train_config}. All results are averaged over the same evaluation protocol to control for stochasticity in sampling-based metrics.

\subsection{Hyperparameters}
\label{sec:hyperparams}

\noindent
\textbf{Notation.} In \autoref{tab:train_config}, \texttt{data\_train\_batch\_size} denotes the \emph{effective} number of sequences per optimizer step after gradient accumulation (i.e., the global batch size across 32 GPUs). \texttt{ppo\_mini\_batch\_size} specifies the per-update minibatch for PPO-style objectives. The \texttt{kl} column is the coefficient of the KL regularizer (set to $0$ for fully KL-free variants). \texttt{length} is the maximum sequence length (tokens) and is kept identical during training and evaluation to avoid truncation bias. \texttt{eval\_step} is the validation interval (in optimizer steps). Method-specific switches are listed under \texttt{Others}.

\begin{table*}[htbp]
    \centering
    \huge
    \renewcommand{\arraystretch}{1.2}
    \resizebox{0.96\textwidth}{!}{
    \begin{tabular}{lcccccccc}
        \toprule
        Method & data\_train\_batch\_size & ppo\_mini\_batch\_size & kl & length & lr & epoch & eval\_step & Others \\
        \midrule
        GRPO  & 128 & 32 & 0.0 & 8k & 1e-6 & 30 & 20 & -- \\
        w/Clip-Higher  & 128 & 32 & 0.0   & 8k & 1e-6 & 30 & 20 & $c_l{=}0.2, c_h{=}0.28$ \\
        w/Ent-Adv  & 128 & 32 & 0.0 & 8k & 1e-6 & 30 & 20 & $\alpha{=}0.4, \kappa{=}2.0$ \\
        w/KL-Cov  & 128 & 32 & 0.0 & 8k & 1e-6 & 30 & 20 & $\lambda_{kl}{=}1.0, \rho_{kl}{=}0.002$ \\
        w/Clip-Cov  & 128 & 32 & 0.0 & 8k & 1e-6 & 30 & 20 & $c_l{=}1.0, c_h{=}1.0, \rho_{clip}{=}0.0002$ \\
        w/AER & 128 & 32 & 0.0 & 8k & 1e-6 & 30 & 20 & $\tau{=}0.4, \rho{=}0.2, \eta{=}0.005$ \\
        \bottomrule
    \end{tabular}
    }
    \caption{\textbf{Training configurations.} We keep the compute budget and decoding policy matched across methods. The KL coefficient controls regularization strength in PPO-style objectives; DAPO removes KL entirely. Clip-Higher uses asymmetric clipping ratios ($c_l$, $c_h$). Ent-Adv introduces entropy-based advantage estimation with scaling factor $\alpha$ and temperature $\kappa$. KL-Cov and Clip-Cov add covariance regularization with coefficients $\rho_{kl}$ and $\rho_{clip}$ respectively. AER (Adaptive Entropy Regulation) introduces adaptive entropy control with target entropy $\tau$, difficulty threshold $\rho$, and adaptation rate $\eta$.}
    \label{tab:train_config}
\end{table*}

\noindent
\textbf{Method-specific parameters:}
\begin{itemize}
    \item \textbf{Clip-Higher:} Uses asymmetric clipping ratios with $c_l = 0.2$ (lower bound) and $c_h = 0.28$ (upper bound) to allow more aggressive updates for positive advantages while constraining negative ones.
    
    \item \textbf{Ent-Adv:} Implements entropy-based advantage estimation with scaling factor $\alpha = 0.4$ and temperature parameter $\kappa = 2.0$ to balance exploration and exploitation.
    
    \item \textbf{KL-Cov:} Adds KL divergence covariance regularization with KL coefficient $\lambda_{kl} = 1.0$ and covariance ratio $\rho_{kl} = 0.002$ to stabilize training dynamics.
    
    \item \textbf{Clip-Cov:} Combines clipping with covariance regularization using symmetric clipping ratios $c_l = c_h = 1.0$ and covariance ratio $\rho_{clip} = 0.0002$ for enhanced stability.
    
    \item \textbf{AER:} Implements adaptive entropy regulation with $\tau = 0.4$, difficulty threshold $\rho = 0.2$ for distinguishing hard vs. easy samples, and adaptation rate $\eta = 0.005$ for dynamic parameter adjustment. The method automatically maintains the target entropy level through adaptive $\alpha$ parameter updates.
\end{itemize}

\paragraph{Scheduling and optimization.}
Unless stated otherwise, we use AdamW with learning rate $1{\times}10^{-6}$ (see \autoref{tab:train_config}) and align the number of epochs so that the \emph{total} number of optimizer updates is comparable across baselines. Evaluation is triggered every \texttt{eval\_step} updates to monitor both pass@\,$1$ (accuracy) and pass@\,$k$ (diversity under multi-sample decoding). For RL methods, rollout budgets (number of samples per prompt) are matched at training and validation time to ensure apples-to-apples comparisons.

\paragraph{Protocol fairness.}
All ablations modify a \emph{single} factor at a time (e.g., toggling KL, changing $\tau$ in AER) while keeping the data curriculum, tokenizer, context length, and decoding policy fixed. This controls confounders and isolates the effect of exploration--exploitation regularization on pass@\,$1$ and pass@\,$k$.

\paragraph{Checkpoint selection.}
For each method, we perform validation on the \textsc{AIME2025} set every 20 optimizer steps (i.e., \texttt{eval\_step}{=}20) using the same decoding configuration as at test time. Unless otherwise stated, the checkpoint used for reporting is the one that attains the highest pass@\,$1$ on \textsc{AIME2025}; all metrics in the main tables are computed from this selected checkpoint.

\section{Detailed Description of Benchmarks}
\label{app:benchmarks}
% 以下是详细的Benchmark的描述。

To fairly evaluate mathematical reasoning ability, we need to use benchmarks that cover different types of problems, various levels of difficulty, and a range of math topics. When choosing datasets, we focus on the following points in Table~\ref{tab:3line-table}:

\begin{table*}[htbp]
\centering
% \large % 使用稍大的字号
\renewcommand{\arraystretch}{2} % 增加行高

\resizebox{0.76\textwidth}{!}{

\begin{tabular}{@{} l p{5cm} p{7cm} @{}}
\toprule
\textbf{Dataset} & \textbf{Core Description} & \textbf{Key Characteristics} \\
\midrule
AIME '24& 
High school Olympiad-level assessment from American Invitational Mathematics Examination & 
\begin{itemize}[leftmargin=*,nosep,itemsep=2pt]
\item 15 complex competition problems
\item Algebra/Geometry/Number theory focus
\item 3-hour time constraint design
\item Multi-step reasoning verification
\end{itemize} \\
\addlinespace[6pt]

\midrule
AIME '25& 
High school Olympiad-level assessment from American Invitational Mathematics Examination & 
\begin{itemize}[leftmargin=*,nosep,itemsep=2pt]
\item 15 complex competition problems
\item Algebra/Geometry/Number theory focus
\item 3-hour time constraint design
\item Multi-step reasoning verification
\end{itemize} \\
\addlinespace[6pt]

\cmidrule(r){1-3}
GSM8K& 
Elementary school math word problem benchmark & 
\begin{itemize}[leftmargin=*,nosep,itemsep=2pt]
\item 8,500 graded problems
\item Natural language scenarios
\item Basic arithmetic operations
\item Step-by-step solution validation
\end{itemize} \\
\addlinespace[6pt]

\cmidrule(r){1-3}
MATH-500& 
Advanced mathematics evaluation set by OpenAI & 
\begin{itemize}[leftmargin=*,nosep,itemsep=2pt]
\item 500 curated problems
\item Formal mathematical notation
\item Non-standard solution analysis
\item Cross-domain evaluation
\end{itemize} \\
\addlinespace[6pt]

\cmidrule(r){1-3}
AMC 2023& 
American Mathematics Competitions system & 
\begin{itemize}[leftmargin=*,nosep,itemsep=2pt]
\item Tiered assessment structure
\item Hybrid question types
\item Curriculum alignment verification
\item Official difficulty metrics
\end{itemize} \\
\bottomrule
\end{tabular}
}
\caption{Comparison of Mathematical Competition Datasets}

\textbf{Links:} \\ 
AIME '24: \url{https://huggingface.co/datasets/HuggingFaceH4/aime_2024};\\ 
AIME '25: \url{https://huggingface.co/datasets/HuggingFaceH4/aime_2025};\\ 
GSM8K: \url{https://huggingface.co/datasets/openai/gsm8k};\\ 
MATH-500: \url{https://huggingface.co/datasets/HuggingFaceH4/MATH-500};\\ 
AMC 2023: \url{https://huggingface.co/datasets/AI-MO/aimo-validation-amc}
\label{tab:3line-table}
\end{table*}

% \clearpage

\section{Additional Experiments and Detailed Results}
\label{app:detailed-results}

Due to space constraints in the main text, we present supplementary experimental results and detailed data in this appendix. We focus on three key aspects: (1) the scalability of our method to larger parameter models, (2) the generalization of our approach to the LLaMA architecture, which has been noted by the community as challenging for mathematical reasoning alignment, and (3) the detailed numerical results corresponding to the experiments presented in the main text. Constrained by computational resources, for these supplementary evaluations, we restricted our comparison to the most widely recognized baselines, and the training duration was limited to fewer than 500 steps.

\subsection{Scalability Analysis (Qwen3-30B-A3B)}
We verified the effectiveness of Adaptive Entropy Regularization (AER) on \textbf{Qwen3-30B-A3B} using the same RLVR configuration. As shown in the top section of Table~\ref{tab:appendix_res}, AER consistently outperforms all baselines across the evaluated benchmarks. Notably, on \textsc{AIME2024}, AER achieves a Pass@32 of 61.6\%, significantly surpassing the Base model (54.2\%) and GRPO (52.5\%). Similarly, on \textsc{MATH500}, it reaches a high accuracy of 93.7\%. These results confirm that the benefits of entropy-aware exploration scale effectively to the 30B parameter class.

\begin{table*}[t]
\centering
\renewcommand{\arraystretch}{1.11}
\scriptsize
\begin{adjustbox}{width=\textwidth}
\begin{tabular}{
  l
  *{8}{S[table-format=2.1]}
}
\toprule
\multicolumn{1}{c}{\multirow[c]{2}{*}{\textbf{Method}}} 
  & \multicolumn{2}{c}{\textbf{AIME24}}
  & \multicolumn{2}{c}{\textbf{AIME25}}
  & \multicolumn{2}{c}{\textbf{MATH500}}
  & \multicolumn{2}{c}{\textbf{AMC23}} \\
\cmidrule(lr){2-3}\cmidrule(lr){4-5}\cmidrule(lr){6-7}\cmidrule(lr){8-9}
& {\it mean@32} & {\it pass@32}
& {\it mean@32} & {\it pass@32}
& {\it mean@8} & {\it pass@8}
& {\it mean@8} & {\it pass@8} \\
\midrule
\multicolumn{9}{c}{\textit{Qwen3-30B-A3B}} \\
\midrule
Base            & 17.6 & 54.2 & 5.8  & 27.8 & 63.3 & 88.7 & 41.2 & 74.3 \\
GRPO            & 25.7 & 52.5 & 15.4 & 37.0 & 87.1 & 92.9 & 68.7 & 87.1 \\
w/ Clip-Higher  & 30.3 & 51.6 & 16.5 & 32.2 & 86.5 & 92.1 & 68.4 & 82.0 \\
\rowcolor{gray!18}
w/ AER (Ours)   & \bfseries 33.0 & \bfseries 61.6 & \bfseries 20.4 & \bfseries 39.6 & \bfseries 88.7 & \bfseries 93.7 & \bfseries 71.2 & \bfseries 90.7 \\
\midrule
\multicolumn{9}{c}{\textit{Llama-3.1-8B-Base}} \\
\midrule
Base            & 0.0 & 1.9 & 0.1 & 2.2 & 9.2  & 30.2 & 2.8 & 13.8 \\
GRPO            & 0.3 & 5.0 & 0.1 & 2.1 & 13.3 & 27.6 & 2.8 & 11.3 \\
w/ Clip-Higher  & 0.4 & 5.3 & 0.1 & 2.0 & 15.9 & 28.8 & 0.9 & 3.9  \\
\rowcolor{gray!18}
w/ AER (Ours)   & \bfseries 0.9 & \bfseries 5.4 & \bfseries 0.3 & \bfseries 4.9 & \bfseries 16.1 & \bfseries 32.9 & \bfseries 2.8 & \bfseries 14.8 \\
\bottomrule
\end{tabular}
\end{adjustbox}
\caption{\textbf{Scalability and Generalization Analysis.} We report the accuracy metrics on \textbf{Qwen3-30B-A3B} and \textbf{Llama-3.1-8B-Base}. Despite the challenging nature of the LLaMA architecture for reasoning tasks, AER consistently outperforms baselines across both model scales and families.}
\label{tab:appendix_res}
\end{table*}

\subsection{Generalization to Different Architectures (Llama-3.1-8B)}
To assess architectural robustness, we evaluated AER on \textbf{Llama-3.1-8B-Base}. Consistent with recent community findings, we observed that the LLaMA family is generally more difficult to train for long-CoT reasoning compared to Qwen, resulting in lower absolute baselines. However, as shown in the bottom section of Table~\ref{tab:appendix_res}, AER still yields consistent relative gains. For instance, on \textsc{MATH500}, AER improves performance to 32.9\% (Pass@8), outperforming both GRPO (27.6\%) and Clip-Higher (28.8\%). This suggests that our method's mechanism is not specific to the Qwen architecture and remains effective in harder-to-align scenarios.

% ---------- AIME2024 ----------
\begin{table*}[t]
  \centering
  \scriptsize
  \renewcommand{\arraystretch}{1.04}
  \begingroup\setlength{\tabcolsep}{8.0pt}
  \resizebox{0.95\textwidth}{!}{
  \begin{tabular}{lrrrrrrrrr}
    \toprule
    \textbf{Method} & \multicolumn{9}{c}{\textbf{AIME2024}: Pass@$k$ (\%)} \\
    \cmidrule(lr){2-10}
     & $k${=}1 & $k${=}2 & $k${=}4 & $k${=}8 & $k${=}16 & $k${=}32 & $k${=}64 & $k${=}128 & $k${=}256 \\
    \midrule
    Base          & 11.5 & 15.5 & 24.2 & 32.1 & 40.2 & 48.0 & 55.1 & 60.9 & 64.3 \\
    GRPO          & 20.5 & 26.2 & 31.3 & 36.4 & 41.5 & 47.0 & 53.5 & 59.5 & 63.7 \\
    w/Clip-Higher & 27.1 & 34.0 & 41.0 & 48.7 & 55.5 & 61.5 & 65.7 & 68.2 & 69.5 \\
    w/AER         & \textbf{31.4} & \textbf{38.2} & \textbf{45.4} & \textbf{52.0} & \textbf{58.3} & \textbf{63.2} & \textbf{66.4} & \textbf{69.4} & \textbf{72.4} \\
    \bottomrule
  \end{tabular}}
  \endgroup
  \caption{\textbf{Qwen3-8B-Base on \textsc{AIME2024}}. Exact Pass@\,$k$ values underlying the main-text curve.}
  \label{tab:passk_aime2024}
\end{table*}

% ---------- AIME2025 ----------
\begin{table*}[t]
  \centering
  \scriptsize
  \renewcommand{\arraystretch}{1.04}
  \begingroup\setlength{\tabcolsep}{8.0pt}
  \resizebox{0.95\textwidth}{!}{
  \begin{tabular}{lrrrrrrrrr}
    \toprule
    \textbf{Method} & \multicolumn{9}{c}{\textbf{AIME2025}: Pass@$k$ (\%)} \\
    \cmidrule(lr){2-10}
     & $k${=}1 & $k${=}2 & $k${=}4 & $k${=}8 & $k${=}16 & $k${=}32 & $k${=}64 & $k${=}128 & $k${=}256 \\
    \midrule
    Base          & 8.8 & 14.6 & 18.9 & 24.5 & 31.0 & 35.5 & 41.5 & 46.7 & 50.3 \\
    GRPO          & 18.5 & 22.5 & 25.7 & 28.4 & 31.1 & 34.3 & 38.1 & 42.5 & 47.1 \\
    w/Clip-Higher & 21.1 & 24.8 & 28.8 & 35.4 & 42.3 & 46.9 & 48.2 & 51.7 & 54.6 \\
    w/AER         & \textbf{25.1} & \textbf{30.1} & \textbf{35.7} & \textbf{41.0} & \textbf{46.0} & \textbf{48.9} & \textbf{54.7} & \textbf{59.4} & \textbf{62.8} \\
    \bottomrule
  \end{tabular}}
  \endgroup
  \caption{\textbf{Qwen3-8B-Base on \textsc{AIME2025}}. Exact Pass@\,$k$ values underlying the main-text curve.}
  \label{tab:passk_aime2025}
\end{table*}

\begin{table}[t]
  \centering
  \small
  \renewcommand{\arraystretch}{1.04}
  \begingroup\setlength{\tabcolsep}{4.0pt}
  \resizebox{1.0\columnwidth}{!}{
  \begin{tabular}{lrrrrr}
    \toprule
    \textbf{AER (\(\rho{=}0.2\))} & \textbf{AIME24} & \textbf{AIME25} & \textbf{AMC23} & \textbf{MATH500} & \textbf{Avg.} \\
    \midrule
    \(\tau{=}0.2\) & 30.8 & 23.8 & 73.7 & 88.9 & 54.3 \\
    \(\tau{=}0.3\) & 29.8 & 23.2 & 75.2 & 88.8 & 54.2 \\
    \(\tau{=}0.4\) & \textbf{31.4} & \textbf{25.1} & 75.6 & \textbf{89.4} & \textbf{55.4} \\
    \(\tau{=}0.5\) & 30.5 & 24.0 & 75.6 & 89.3 & 54.8 \\
    \(\tau{=}0.6\) & \textbf{31.4} & 23.2 & 75.9 & 89.2 & 54.9 \\
    \(\tau{=}0.8\) & 30.0 & 24.1 & \textbf{76.6} & 88.9 & 54.9 \\
    \(\tau{=}1.0\) & 31.1 & 23.2 & 72.8 & 89.2 & 54.1 \\
    \(\tau{=}1.2\) & 28.7 & 20.7 & 74.2 & 88.8 & 53.1 \\
    \bottomrule
  \end{tabular}}
  \endgroup
  \caption{\textbf{AER target-entropy ablation (\(\rho{=}0.2\)).} Exact accuracy (\%). \textbf{Avg.} is the mean over the four datasets. Best values per column are in bold.}
  \label{tab:aer_tau_ablation}
\end{table}

\begin{table}[t]
  \centering
  \small
  \renewcommand{\arraystretch}{1.04}
  \begingroup\setlength{\tabcolsep}{4.0pt}
  \resizebox{1.0\columnwidth}{!}{
  \begin{tabular}{lrrrrr}
    \toprule
    \textbf{AER (\(\tau{=}0.4\))} & \textbf{AIME24} & \textbf{AIME25} & \textbf{AMC23} & \textbf{MATH500} & \textbf{Avg.} \\
    \midrule
    \(\rho{=}0.0\) & \textbf{33.1} & 24.6 & 75.6 & \textbf{89.7} & \textbf{55.8} \\
    \(\rho{=}0.2\) & 31.4 & 25.1 & 75.6 & 89.4 & 55.4 \\
    \(\rho{=}0.4\) & 30.2 & \textbf{26.1} & \textbf{76.4} & 89.3 & 55.5 \\
    \(\rho{=}0.6\) & 32.1 & 23.7 & 75.1 & 89.5 & 55.1 \\
    \(\rho{=}0.8\) & 29.5 & \textbf{26.1} & 75.9 & 89.4 & 55.2 \\
    \(\rho{=}1.0\) & 29.4 & 21.4 & 73.8 & 89.4 & 53.5 \\
    \bottomrule
  \end{tabular}}
  \endgroup
  \caption{\textbf{AER ablation on the difficulty threshold \(\rho\) (fixed \(\tau{=}0.4\)).} Exact accuracy (\%). \textbf{Avg.} is the mean over the four datasets. Best values per column are in bold.}
  \label{tab:aer_rho_ablation}
\end{table}

\subsection{Pass@\texorpdfstring{$k$}{k} as a Function of \texorpdfstring{$k$}{k} (Qwen3\,-\,8B\,-\,Base)}
\label{app:pass_at_k}
Table~\ref{tab:passk_aime2024}--\ref{tab:passk_aime2025} report the exact values underlying the Pass@$k$--vs.--$k$ plots on \textsc{AIME2024} and \textsc{AIME2025}. We follow the uniform evaluation setup and the checkpoint-selection protocol in the main text (validation every 20 steps on \textsc{AIME2025} and selecting the checkpoint with the highest Pass@\,$1$ per method). All numbers are percentages and are monotone non-decreasing in $k$ as expected.

\paragraph{Statistical remarks.}
(1) \textbf{Small-$k$ accuracy.} At $k{=}32$, \emph{w/AER} attains 63.2\% on \textsc{AIME2024} and 48.9\% on \textsc{AIME2025}, improving over \emph{Base} (48.0\%, 35.5\%) by {+15.2}/{+13.4} points and over \emph{GRPO} (47.0\%, 34.3\%) by {+16.2}/{+14.6} points; it also surpasses \emph{w/Clip-Higher} (61.5\%, 46.9\%) by {+1.7}/{+2.0} points.  
(2) \textbf{Exploration headroom.} The gain from $k{=}32$ to $k{=}256$ for \emph{w/AER} is {+9.2} points on \textsc{AIME2024} (63.2$\rightarrow$72.4) and {+13.9} on \textsc{AIME2025} (48.9$\rightarrow$62.8), exceeding \emph{w/Clip-Higher} on both splits ({+8.0}, {+7.7}), indicating a larger pool of viable solutions at higher $k$.  
(3) \textbf{Sample efficiency via \(k_{50}\).} Define \(k_{50}=\min\{k:\mathrm{Pass@}k\ge 50\%\}\). On \textsc{AIME2024}, \(k_{50}\) is 8 for \emph{w/AER}, 16 for \emph{w/Clip-Higher}, and 64 for both \emph{Base} and \emph{GRPO}. On \textsc{AIME2025}, \(k_{50}\) is 64 for \emph{w/AER}, 128 for \emph{w/Clip-Higher}, 256 for \emph{Base}, while \emph{GRPO} does not reach 50\% by \(k{=}256\). These trends are consistent with the intended effect of adaptive entropy regularization.

\subsection{Ablation on Reduction Ratio \texorpdfstring{\(\tau\)}{tau} (AER, \(\rho{=}0.2\))}
\label{app:ablation_tau}

We ablate the target-entropy ratio \(\tau\in\{0.2,0.3,0.4,0.5,0.6,0.8,1.0,1.2\}\) with \(\rho\) fixed at \(0.2\) on Qwen3-8B-Base. 
Results in Table~\ref{tab:aer_tau_ablation} show that overall accuracy peaks near \(\tau{\approx}0.4\) (\(\text{Avg.}=55.4\)), and degrades as \(\tau\) increases into the high-entropy regime (\(\tau{\ge}1.0\): \(54.1/53.1\)). 
On the hard splits (\textsc{AIME2024/2025}), moderate entropy reduction improves accuracy up to \(\tau{=}0.4\) (\(31.4/25.1\)), whereas pushing \(\tau\) higher reduces performance on \textsc{AIME2025}. 
On \textsc{AMC23} and \textsc{MATH500}, the trend is milder; \textsc{AMC23} peaks at \(\tau{=}0.8\) (76.6), but the overall average remains best at \(\tau{=}0.4\). 
These observations support the existence of a favorable entropy band centered around \(\tau\approx0.4\).

\subsection{Ablation on Difficulty Threshold \texorpdfstring{\(\rho\)}{rho} (AER, \(\tau{=}0.4\))}
\label{app:ablation_rho}

We ablate the difficulty threshold \(\rho\) (formerly denoted \(\rho\); cf.\ Section~\S\ref{sec:aer}) while fixing the target-entropy ratio at \(\tau{=}0.4\). Recall that samples with group accuracy \(g(q)\le\rho\) receive a positive entropy bonus (hard set). 
Table~\ref{tab:aer_rho_ablation} shows that \textbf{narrower hard sets} (smaller \(\rho\)) generally yield higher overall accuracy: the best average is obtained at \(\rho{=}0.0\) (55.8), and performance degrades when \(\rho\) is made overly inclusive (e.g., \(\rho{=}1.0\), 53.5). 
This indicates that allocating the entropy budget only to the \emph{genuinely} difficult prompts preserves exploitation on easier ones and avoids diluting exploration across too many samples. 
At the dataset level, \textsc{AIME2025} and \textsc{AMC23} achieve their peaks at moderate \(\rho\) (0.4/0.8 and 0.4, respectively), suggesting a mild benefit from targeted—but not indiscriminate—expansion of the hard set.

\end{document}